\documentclass[lettersize,journal]{IEEEtran}
\usepackage{amsmath,amsfonts,amssymb}
\usepackage{algorithmic}
\usepackage{array}
\usepackage[caption=false,font=normalsize,labelfont=sf,textfont=sf]{subfig}
\usepackage{textcomp}
\usepackage{stfloats}
\usepackage{url}
\usepackage{verbatim}
\usepackage[shortlabels]{enumitem}
\usepackage{graphicx}
\usepackage{adjustbox}
\usepackage{xcolor}%
\usepackage{colortbl}
\usepackage{booktabs}
\usepackage{placeins}

\makeatletter
\def\@begintheorem#1#2{\par\addvspace{8pt plus3pt minus2pt}%
\noindent{\bfseries #1\ #2.\ }\itshape}
\makeatother

\newtheorem{theorem}{\textbf{Theorem}}[section]
\newtheorem{proposition}[theorem]{\textbf{Proposition}}
\newtheorem{lemma}[theorem]{\textbf{Lemma}}

\def\BibTeX{{\rm B\kern-.05em{\sc i\kern-.025em b}\kern-.08em
    T\kern-.1667em\lower.7ex\hbox{E}\kern-.125emX}}
\usepackage{balance}
\begin{document}
\title{Realigned Softmax Warping for Deep Metric Learning}
\author{Michael G. DeMoor, \textit{Student Member}, IEEE and John J. Prevost, \textit{Senior Member} IEEE \thanks{This work has been submitted to the IEEE for possible publication. Copyright may be transferred without notice, after which this version may no longer be accessible.}}

\markboth{Manuscript}
{How to Use the IEEEtran \LaTeX \ Templates}

\maketitle

\begin{abstract}
    Deep Metric Learning (DML) loss functions traditionally aim to control the forces of separability and compactness within an embedding space so that the same class data points are pulled together and different class ones are pushed apart. Within the context of DML, a softmax operation will typically normalize distances into a probability for optimization, thus coupling all the push/pull forces together. This paper proposes a potential new class of loss functions that operate within a euclidean domain and aim to take full advantage of the coupled forces governing embedding space formation under a softmax. These forces of compactness and separability can be boosted or mitigated within controlled locations at will by using a warping function. In this work, we provide a simple example of a warping function and use it to achieve competitive, state-of-the-art results on various metric learning benchmarks.
\end{abstract}


\section{Introduction}
Metric Learning is an influential field that has many applications in different sub-domains of AI and Computer Vision. It consists of using data to learn a distance metric for measuring similarity/dissimilarity between samples. Deep Metric Learning (DML) combines Metric Learning with Deep Learning to further enhance performance. This approach has been successfully used to improve applications in areas like Information Retrieval/Visual Search 
\cite{movshovitz2017no,teh2020proxynca++,kim2020proxy,sun2020circle}, Face Recognition/Verification \cite{deng2019arcface,wang2018cosface,sun2020circle}, Person Re-Identification \cite{liao2022graph,chen2017beyond}, and Image Segmentation \cite{fathi2017semantic,neven2019instance}.

\begin{figure}[t]
    \centering
    \subfloat[Normal Approach]{
        \centering
        \includegraphics[width=0.75\columnwidth, height=3.25cm]{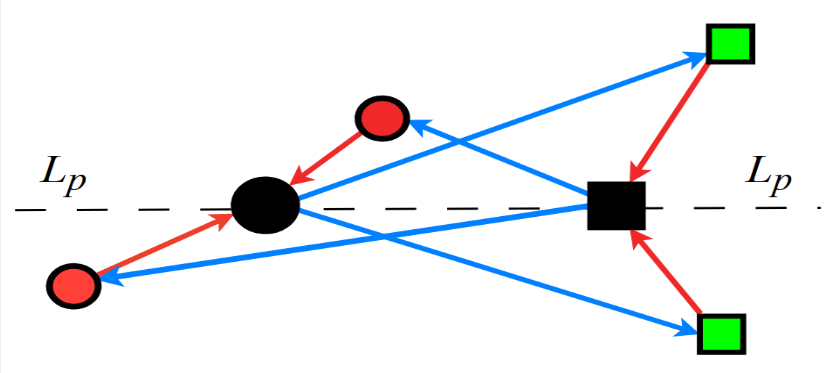}
        \label{fig:Loss-Comparison-Normal}
    }\\
    \subfloat[Realigned Approach]{
        \centering
        \includegraphics[width=0.85\columnwidth, height=3.25cm]{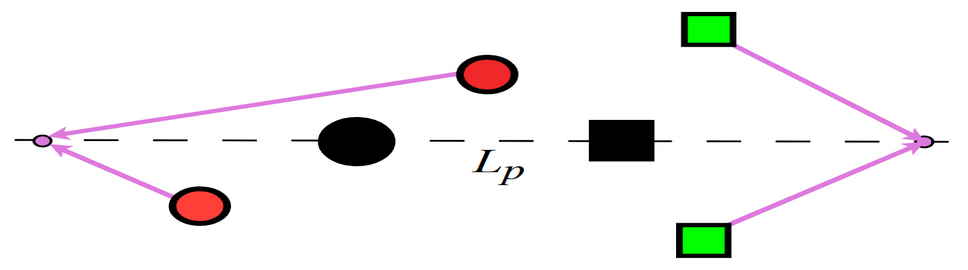}
        \label{fig:Loss-Comparison-Realigned}    
    }
    \caption{
    Binary-Class diagram comparing the forces of traditional proxy-losses vs. our approach. 
    Best viewed in color. Each node is an embedding vector. Different shapes are different classes. 
    Red/blue/magenta lines indicate pull/push/coupled forces respectively. 
    Black nodes are proxies. 
    a) Traditional proxy losses are governed by interacting push/pull forces that could negatively interfere with each other. 
    b) Instead of pulling embeddings toward their proxies and pushing them away from other proxies, our loss deals with the coupled minimums of both forces directly and realigns them outward. This encourages embeddings to move away from \emph{both} proxies (including their own) 
    towards a single outward point which boosts separability.
    }
    \label{fig:Proxy-Loss-Comparison}
\end{figure}


In DML, a deep network (e.g. a convolutional neural network) is used to map data samples into an embedding space. A loss function is then applied in order to encourage the model to learn to group together similar embedding vectors (e.g. those belonging to the same class) and to separate dissimilar ones. Improvements in the field of DML can be roughly categorized according to the method of improvement. Some methods (e.g. \cite{ebrahimpour2022multi}) focus on improving the design of the overall network architecture used to learn the similarity metric. Other works (\cite{elezi2020group,sanakoyeu2019divide,wu2017sampling}) will enhance the training methods and techniques used to train these models. Still others (\cite{sohn2016improved,oh2016deep,sun2020circle,teh2020proxynca++,kim2020proxy} and this work) prioritize the design of more potent loss functions. Some traditional loss function examples include the Contrastive \cite{hadsell2006dimensionality} and Triplet \cite{schroff2015facenet} losses. These can be combined with a specialized sampler or training method (e.g.\cite{harwood2017smart}) to make them more effective. 

The Softmax operation is commonly used as a method of normalizing distances into a probability that can be forwarded through a loss function. 
All DML loss functions aim to control the push/pull forces responsible for embedding space formation. However, the standard softmax operation can complicate matters by coupling them together. This makes it more difficult to prioritize which force to enhance/mitigate and when.
There has been some work to dissect the softmax into separate forces of compactness and separability \cite{he2020softmax}. 
We take a different approach and opt to exploit the fully coupled natural forces instead.
There are methods that employ a softmax operation in their designs \cite{zhai2018classification,qian2019softtriple, teh2020proxynca++, sun2020circle, boudiaf2020unifying, yang2023stop, wang2018cosface, deng2019arcface, liu2016large, ranjan2017l2}. However, they're typically built around hyperspherical embedding spaces and cosine distances. 
In this work, we investigate a potential alternative way to formulate new loss functions that utilize an unbounded space.
As such, we adopt the euclidean metric and formally explore the theory behind the complex coupled interacting forces at work under a softmax. 
We then use this theory to modify the vanilla softmax to boost separability/compactness and create superior/competitive euclidean embedding spaces.
In summary, we contribute the following:
\begin{itemize}
    \item We break down the natural softmax and dissect/explore the push/pull forces at work within a euclidean space in greater detail than has been done before.
    \item We introduce some basic theory that can be used to develop a new class of loss functions and offer a simple example.
    \item Using this example, we achieve competitive, state-of-the-art results on standard metric learning benchmarks \cite{welinder2010caltech,krause20133d,oh2016deep}.
\end{itemize}

\section{Preliminaries}
\subsection{Loss Types}
\label{subsec:Loss-Types}

DML loss functions can be largely categorized into two primary types. 
Traditional pair-based losses (e.g. \cite{weinberger2005distance, wang2019multi, hadsell2006dimensionality}) will organize data samples into positive and negative tuples/pairs and optimize the model to position positives closer towards each other and further away from negatives.
The Contrastive and Triplet losses are classic pair-wise examples. Contrastive loss simply pulls together samples of the same class and pushes apart differing ones. Triplet loss takes a sample as an anchor and compares it with both a positive sample and a negative sample. The positive pair distance is then constrained to be less than that of the negative pair. There have been various advancements ranging from generalizing the number of negative pairs per-anchor \cite{oh2016deep, sohn2016improved} to enhancing sampling strategies and constructing pair-based losses around them \cite{wang2019multi, wu2017sampling} and more (e.g. \cite{kan2022contrastive}). There are also pairwise-ranking methods that take into account the order of the nearest neighbor pairs within a batch \cite{patel2022recall, revaud2019learning, cakir2019deep, ramzi2021robust, yuan2023osap, brown2020smooth, rolinek2020optimizing, liao2023supervised}.

\begin{figure*}[t]
    \centering
    \includegraphics[width=\textwidth, height=3.0cm]{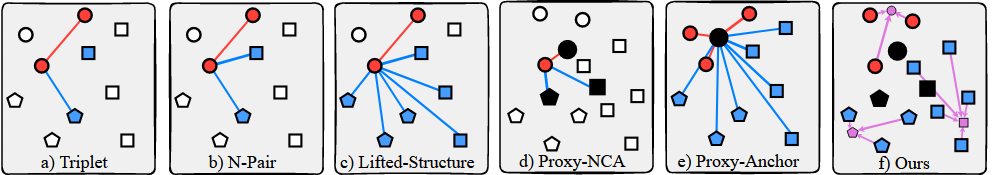}
    \caption{
    Loss comparison. Best viewed in color. Each node is an embedding vector. Different shapes are different classes. Red/blue/magenta lines indicate pull/push/coupled forces respectively. Black nodes are proxies. Small magenta nodes are coupled points of attraction. a) Triplet loss \cite{weinberger2005distance} pulls an anchor closer to a positive sample and pushes away a negative. b)-c) N-Pair \cite{sohn2016improved} and Lifted-Structure \cite{oh2016deep} generalize this further using multiple negatives. d) Proxy-NCA \cite{movshovitz2017no} compares a sample against a proxy for each class. e) Proxy-Anchor \cite{kim2020proxy} associates all data in a batch with each proxy.
    f) Instead of pulling embeddings toward their proxies, our loss realigns coupled push/pull forces toward outward points of attraction during training thus boosting separability. 
    }
    \label{fig:Loss-Comparison}
\end{figure*}

Proxy (or classification)-based losses (e.g. \cite{movshovitz2017no, teh2020proxynca++, kim2020proxy, zhu2020fewer}) utilize an extra layer of weights/proxies and will mitigate training complexity by bypassing the need to organize tuples of pair-wise data altogether. The concept of proxies was introduced in \cite{movshovitz2017no}. They are network parameters (separate from the base model) that ideally learn to represent the training data and are typically (though not always) assigned per-class. 
Traditionally, proxy-based loss functions will encourage each data point to associate more closely with its corresponding proxy and vice-versa for all the other proxies. ProxyNCA \cite{movshovitz2017no} utilizes neighborhood component analysis (NCA) to achieve this. ProxyNCA++ \cite{teh2020proxynca++} improves on ProxyNCA by adjusting the NCA portion to normalize the distances into a probability. 
Proxy-Anchor \cite{kim2020proxy} uses proxies as anchors in order to leverage data-to-data relations normally missing in proxy-based losses. SoftTriple \cite{qian2019softtriple} experiments with multiple proxies per-class, and Hierarchical Proxy \cite{yang2022hierarchical} generalizes arbitrary proxy-based losses by using proxy pyramids to model inherent data hierarchies. ArcFace \cite{deng2019arcface}, CosFace \cite{wang2018cosface}, Large-Margin Softmax \cite{liu2016large}, L2-Softmax \cite{ranjan2017l2} and
can all be categorized as classification/proxy-based losses as well. 
SGSL \cite{yang2023stop} is another more recent loss that selectively applies a scaling term to, in essence, increase same class similarity and decrease otherwise. It too can be categorized as a proxy-based loss.
There are also methods like \cite{seidenschwarz2021learning,lim2022hypergraph} that work in combination with a graph neural network. Our loss builds on top of \cite{teh2020proxynca++} but modifies the coupled forces under the softmax within a euclidean setting to complement separability (see Figs. \ref{fig:Proxy-Loss-Comparison} and \ref{fig:Loss-Comparison}). 

\subsection{Softmax Embeddings}
\label{subsec:Softmax-Embeddings}
Given a mini-batch of N training Samples \begin{math}
\mathbf{X} = \{x_i\}_{i=1}^N 
\end{math}
and their corresponding Labels \begin{math}
\mathbf{Y} = \{y_i\}_{i=1}^N 
\end{math}, of which each belongs to one of \(\mathcal{C}\) classes, in DML, the model will learn a function to map each input \(x_i\) to a corresponding \(n\)-dimensional feature embedding vector \(e_{i}\). 
\begin{equation}
    \phi(\mathbf{X}) = \{e_i\}_{i=1}^N
\end{equation}
Depending on how it is employed, the softmax operation has various implications on how the resulting embedding space is created. 
A modest number 
of the aforementioned losses make use of a softmax. In fact, according to \cite{boudiaf2020unifying}, even pair-based losses are (in effect) approximated via bound-optimization by the vanilla Softmax-Cross Entropy Loss. The traditional Cross Entropy (CE) Loss is written below (for a single sample).
\begin{equation}
CE = -\log\Big(\frac{\exp\big(d(e_i, p_{y_i})\big)}{\sum_{j=1}^\mathcal{C} \exp\big(d(e_i,{p_j})\big)} \Big) \\
\end{equation}
\begin{equation}
 = \;\; \log\Big(1 + {\sum_{j \neq y_i}^\mathcal{C} \exp\big(d_1 - d_2\big)} \Big)
 \label{eq:Default-Softmax}
\end{equation}
where \(d_1 = d(e_i,p_j)\), \(d_2 = d(e_i,p_{y_i})\) and \(p\) represents the weights (proxies) from the final classification layer. Note that in it's current form the CE Loss will encourage \(d_1\) to be small and \(d_2\) large. Loss functions that use this form (e.g. some cosine losses) will encourage a large (similarity) value between \(e_i\) and it's ground truth class \(p_{y_i}\).  

\begin{figure*}[t]
    \subfloat[]{%
        \includegraphics[width=0.24\textwidth, height=4cm]{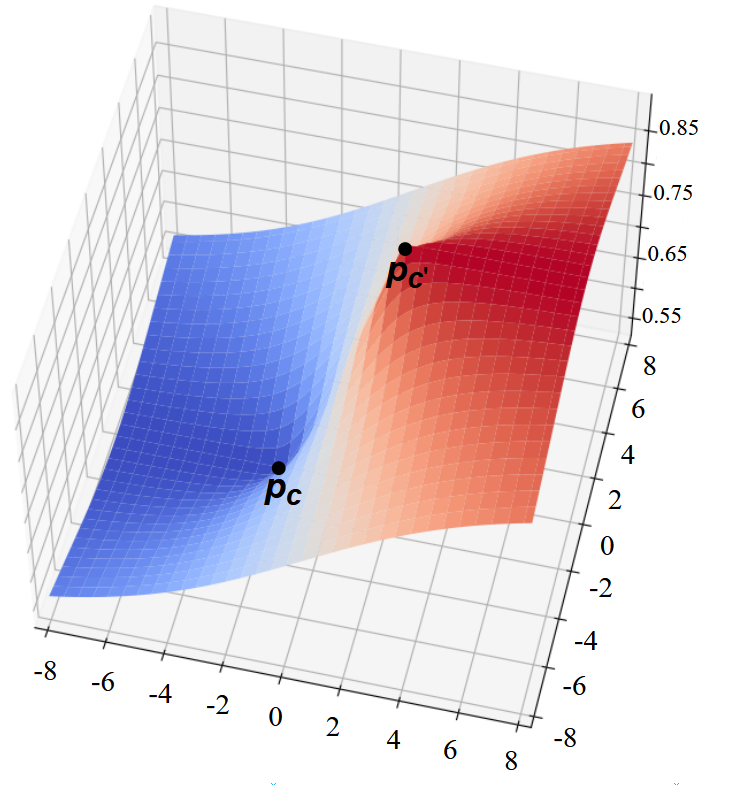}%
        \label{fig:Optimization-Landscape-Normal}%
    }
    \hfill
    \subfloat[]{%
        \includegraphics[width=0.24\textwidth, height=4cm]{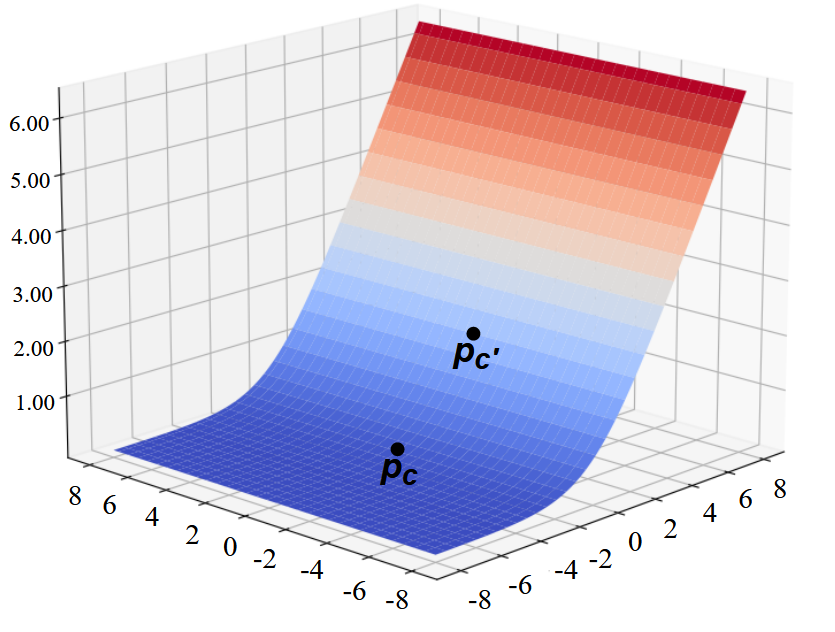}%
        \label{fig:Optimization-Landscape-Convex}%
    }
    \hfill
    \subfloat[]{%
        \includegraphics[width=0.24\textwidth, height=4cm]{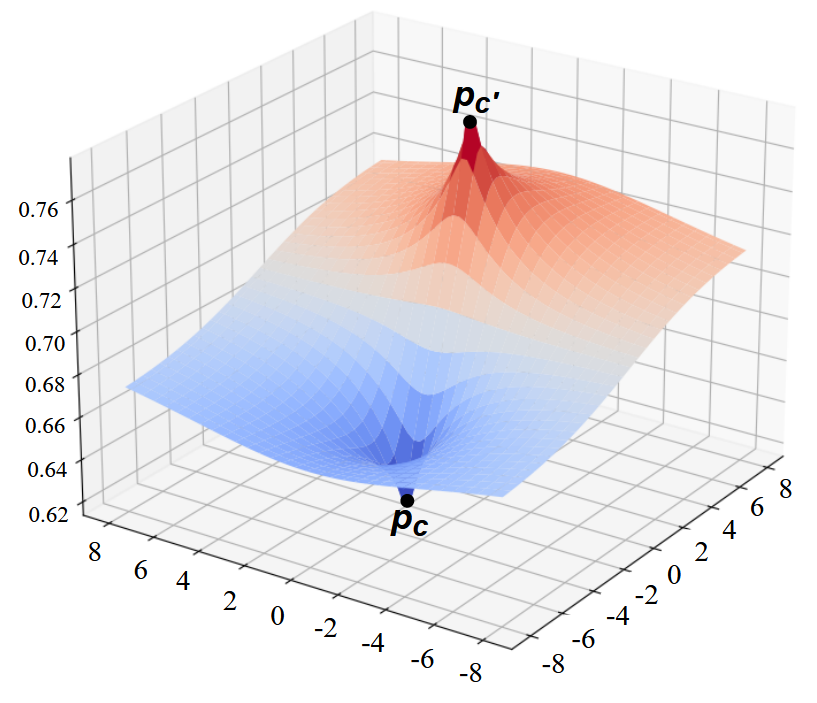}%
        \label{fig:Optimization-Landscape-Concave}%
    }
    \hfill
    \subfloat[]{%
        \includegraphics[width=0.25\textwidth, height=4cm]{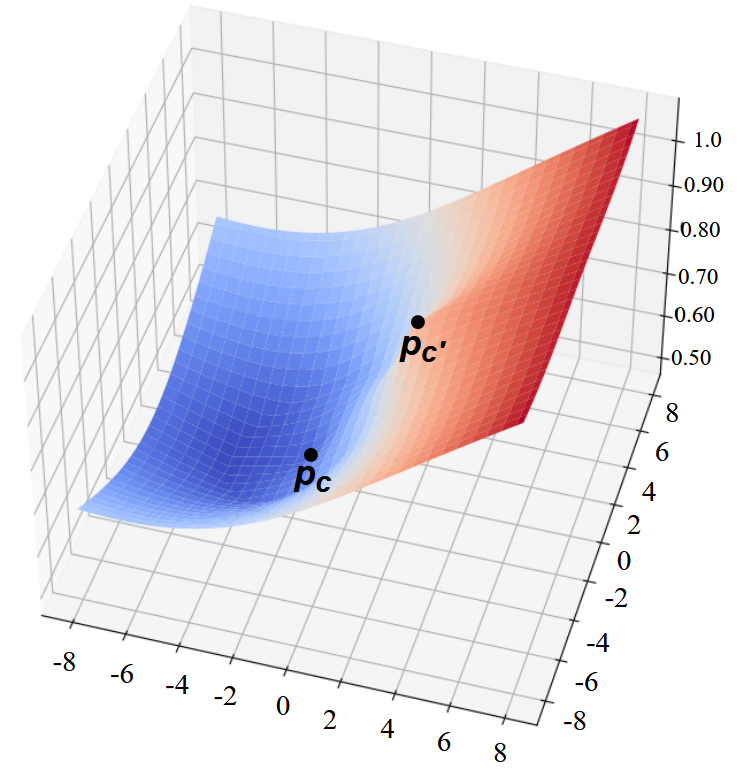}%
        \label{fig:Optimization-Landscape-Warped}%
    }
    \caption{Softmax optimization landscapes for various functions. (a) \(f = t\) (Laplacian Kernel): The default unwarped landscape in Eq. \ref{eq:Default-Softmax-Inverted}. 
    (b) \(f = t^2\) (Gaussian Kernel): This convex function prioritizes separability with no focus on compactness. (c) \(f = \sqrt{t}\): This concave function warps space inwards towards the ground truth proxy, hurting separability. 
    (d) An approximation of a slight warp using \(f_1\) = Eq. \ref{eq:Warp-Function}, \(\alpha = 3.0, k_1 = 0.65, k_2 = 1.5\), \(f_2 = t\): Here the global minimum is slightly shifted outward away from \(p_c\) (boosting separability), and space that is further away is warped inward to preserve compactness. Best viewed in color and zoom.}    
    \label{fig:Optimization-Landscape}
\end{figure*}

Most applications of DML utilize a cosine/inner product (on top of normalized embeddings and proxies) as the `distance'. However, any metric can be used here. 
Regardless, 
minimizing a softmax loss constitutes minimizing a symmetrical \(d_1 - d_2\) term. The gradients for both \(d_1\) and \(d_2\) are effectively equal but opposite (decreasing one distance has the same effect as increasing the other). As explained in \cite{sun2020circle}, the \(d_1 - d_2\) term induces a linear decision boundary which corresponds to many possible points of optimum. Circle Loss approaches the problem by generalizing the \(d_1 - d_2\) term to warp the decision margin into a circle, thus creating a single point of optimum. However, Circle Loss limits its discussion to the scope of the \(d_1 - d_2\) optima only and does not analyze any patterns of local or global minimums within the underlying embedding space. 
The locations of the minimums within the optimization landscape will have different patterns depending on the employed distance metric and embedding manifold.  

\subsection{Cosine vs. Euclidean}
\label{subsec:CosinevsEuclidean}
In this work we choose to study the phenomenon in an unconstrained euclidean embedding space. And we do this for a couple reasons. The first is that the intuition behind our approach (which will be explained in detail in subsequent sections) makes more natural sense within a euclidean setting. Referring to Figs. \ref{fig:Proxy-Loss-Comparison} \& \ref{fig:Loss-Comparison}, our method serves to realign the forces of attraction outward to complement separability. To do so, we select an ``outward" point that strategically serves to attract embeddings outward. This point is more difficult to define within a closed and bounded domain like a hypersphere (where you only have a finite amount of room to expand outward on any given geodesic before encircling the sphere). In a euclidean domain space is unbounded. So this point can be treated as a simple hyperparameter on an infinite line segment. As such, the implementation is simpler. The vast majority of the aforementioned loss functions utilize cosine similarity as their metric of choice. So the other reason for adopting a euclidean base metric is simply to explore possible alternatives. This includes potential benefits in applications where the magnitude of a vector could be better utilized to improve performance (e.g. \cite{meng2021magface}).

\subsection{Definitions}
\label{subsec:Definitions}
We make use of the following terminology and definitions throughout this paper. Given two separate clusters \(C\) and \(C^\prime\) with corresponding proxies \(p_c\) and \(p_{c^{\prime}}\), for \(R \in \mathbb{R}\),
the \(R\)-Disk is defined as:
\begin{center}
    \(D_R(p) := \{e \in \mathbb{R}^n: ||e - p|| = R\}\)
\end{center}
where \(||\cdot||\) denotes the L2-Norm.
We define the line of intersection:
\begin{center}
    \textit{\(L_p :=\) the infinite line intersecting \(p_c\) and \(p_{c^{\prime}}\)}
\end{center}
And we define \(e_*\) and \(e^*\) as the points of \(D_R(p)\) that intersect \(L_p\):
\begin{center}
    \(\{e_*, e^*\} := \{e \in D_R(p) \cap L_p\}\)
\end{center}
Furthermore, we refer to the set of twice smooth and differentiable functions as \(C^2\) and monotone non-decreasing functions as simply monotone.

\section{Realigned Softmax Warping}
Working within an unbounded euclidean space, we adopt the intuition that embeddings should be physically ``close" if they are similar and ``far" apart otherwise. This requires an inversion of the exponential term in Eq. \ref{eq:Default-Softmax} (\(d_1 - d_2\) becomes \(d_2 - d_1\)). 
To that end, we start our proposed method by inverting the exponential term and formally adopting the euclidean distance. Furthermore, as in \cite{teh2020proxynca++}, we assign a single proxy per-class. However, we emphasize that neither embeddings nor proxies are normalized onto a hypersphere.
\begin{equation}
    CE = \log \Big(1 + \sum_{j \neq y_i}^\mathcal{C} \exp({||e_i - p_{y_i}|| - ||e_i - p_j||})\Big)
    \label{eq:Default-Softmax-Inverted}
\end{equation}

The vanilla CE loss in Eq. \ref{eq:Default-Softmax-Inverted} will accomplish the metric learning objective to an extent (see experiments), but it is more likely to suffer from a worse combination of separation and compactness (see Fig. \ref{fig:Optimization-Landscape} and Fig. \ref{fig:Toy-Example-Official}).
The following sections explain in greater detail why this is the case. 
There is a TLDR of this section in Appendix A and simplified explanations/further discussion in Appendix B. We welcome the reader to read through it. 


\subsection{Embedding Formation}
\label{subsec:Embedding-Formatino}
Eq. \ref{eq:Default-Softmax-Inverted} will be minimized through a series of complex coupled interacting forces. 
We simplify our analysis of them by first considering only 2 classes: the correct class \(c = y_i\) and \(c^\prime \neq y_i\). Fig. \ref{fig:Optimization-Landscape-Normal} illustrates its optimization landscape. 
As is, the landscape contains a global minimum at \(p_c\) and then plateaus outward. There is no further incentive for embeddings to separate from the opposite class (\(p_{c^\prime}\)) while remaining compact. 
Under this simplified binary class setting, for a given embedding vector \(e\) there are four possibilities/forces of optimization:

\begin{enumerate}[start=1, label=\textbf{{F\arabic*)}}]
    \item \(e\) can be moved further away from \(p_{c^{\prime}}\)
    \item  \(p_{c^{\prime}}\) can be moved further away from \(e\)
    \item \(e\) can be moved closer to its own proxy \(p_c\)
    \item \(e\)'s ground truth proxy \(p_c\) can be moved closer to \(e\)
\end{enumerate}

\begin{figure}[t]
    \centering
    \includegraphics[width=\columnwidth, height=2.25cm]{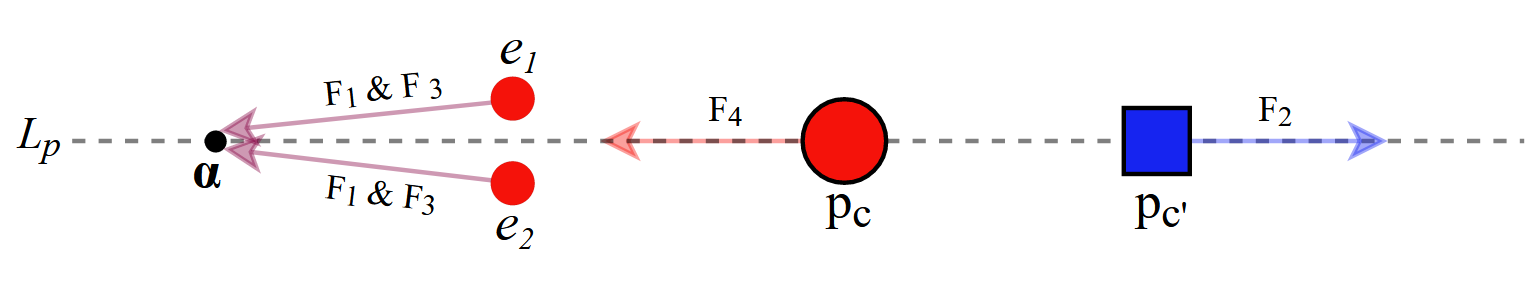} 
    \caption{Intuitively speaking, as training progresses and embeddings group together, the coupled forces acting directly on them will begin to align with the forces acting on the proxies.}
    \label{fig:Aligned-Forces}
\end{figure}

Note that since \(e\) is in both distance terms in Eq. \ref{eq:Default-Softmax-Inverted}, F1 and F3 are coupled together. 
Temporarily shifting focus to these forces, 
there is a minimum that coincides with F1, a minimum that coincides with F3, and a total minimum that culminates from a combination of both. 
If the proxies 
are held constant, this coupled minimum is the most optimal position for \(e\) to settle in order to minimize Eq. \ref{eq:Default-Softmax-Inverted}. 
Ideally, the best place to position a minimum would be at a location that boosts separability and preserves compactness. 
This location
can be explicitly influenced via modifications to the distance terms. 
Consider the following generalization of Eq. \ref{eq:Default-Softmax-Inverted} (again, under the binary setting):
\begin{equation}
     CE = \log \Big(1 + \exp\big({f(||e_i - p_c||) - f(||e_i - p_{c^{\prime}}||)}\big)\Big)
     \label{eq:Function-Softmax}
\end{equation}
where \(f : [0,\infty) \longrightarrow [0,\infty)\) is added to investigate potential modifications to the landscape. 
We now propose a simple yet fundamental fact about \(f\) and its relationship to embedding space formation.

\begin{flushleft}
\begin{lemma}
\textit{Given both \(p_c\) and \(p_{c^\prime}\), \(f\) is monotone iff for any \(r \in \mathbb{R}\),  \(e_*\) and \(e^*\) are respectively the only minimum and maximum of Eq. \ref{eq:Function-Softmax} within \(D_r(p_{c^{\prime}})\).}
\label{lemma:Lp-Extrema}
\end{lemma}
\end{flushleft}
The proof is in Appendix C. This makes clear that constraining \(f\) to monotone functions will place any global extrema of Eq. \ref{eq:Function-Softmax} onto \(L_p\). Furthermore, it implies that regardless of proxy locations there are no local minima embedded throughout the landscape other than \(L_p\). It also makes clear that only monotone functions are capable of this. However, it does not specify where on \(L_p\) any extrema reside.

In order to boost separability while preserving compactness, it is important to control points of attraction. 
Ideally, the best optimum would be one that encourages each cluster to move \emph{directly} away from the other \emph{without} ripping apart.
This would consist of encouraging their respective embedding vectors to move in relative unison away in the opposite direction.  
To achieve this, we propose to place a unique global minimum at a predefined point of attraction \(\alpha\) in-between \(p_c\) and \(\infty\)  
instead of directly at \(\infty\) (perfect separation) or directly at \(p_c\) (perfect compactness).
This also serves to realign F1 and F3 so that the coupled forces act in an \emph{outward} direction (Fig. \ref{fig:Aligned-Forces} and Fig. \ref{fig:Loss-Comparison}).
This is difficult to accomplish utilizing a single \(f\). 
Instead, we generalize Eq. \ref{eq:Function-Softmax} by splitting it up. 
\begin{equation}
     CE = \log \Big(1 + \exp\big(f_1(t_1) - f_2(t_2)\big)\Big)   
     \label{eq:Function-Softmax-Split}
\end{equation}
Where \(f_1\) and \(f_2\) are applied to \(t_1 = ||e - p_c||\) and \(t_2 = ||e - p_{c^\prime}||\) respectively.
We abbreviate the derivatives \(\frac{df_1(t_1)}{dt_1}|_{t_1=||e - p_c||}\) and \(\frac{df_2(t_2)}{dt_2}|_{t_2=||e - p_{c^{\prime}}||}\) as simply \(\frac{df_1}{dt}\) and \(\frac{df_2}{dt}\) respectively.
\begin{flushleft}
\begin{proposition} \textbf{(Simplified):}
\textit{Reversing the \(\frac{df_1}{dt} < \frac{df_2}{dt}\) inequality at any point \(e \in L_p\) is equivalent to creating a minimum in Eq. \ref{eq:Function-Softmax-Split} at \(e\). 
}
\label{prop:Minimum-Creation}
\end{proposition}
\end{flushleft}
The formal statement and proof is in Appendix C. 
According to Lemma \ref{lemma:Lp-Extrema}/Prop. \ref{prop:Minimum-Creation}, the only way to control the location of a unique global minimum is by controlling where the inequality \(\frac{df_1}{dt} < \frac{df_2}{dt}\) changes (with monotone \(f_{1,2}\)). A monotone \(f_{1,2}\)
constrains potential extrema to \(L_p\). Subsequently, sliding a point \(e\) down this line, if \(\frac{df_1}{dt} < \frac{df_2}{dt}\), then \(f_2\) is increasing \emph{faster} than \(f_1\). 
Therefore, \(f_1 - f_2\) is decreasing and \(e\) is going \emph{downhill}. 
Similarly, if \(\frac{df_1}{dt} > \frac{df_2}{dt}\), then \(f_1 - f_2\) is increasing and \(e\) is going \emph{uphill}. By reversing these inequalities at some point \(\alpha\), \(e\) is \emph{switched} from going downhill (decreasing) to going uphill (increasing). That is, a minimum is created at the point of reversal \(\alpha\).

\subsection{Function Warping}
\label{subsec:Function-Warping}
There are two potential extremes when choosing \(f_{1,2}\). The first occurs when \(\frac{df_1}{dt} < \frac{df_2}{dt}\) everywhere. Some example cases where this occur are when \(f_1 = f_2\) and is strictly convex. 
This follows from the definition of convexity and the fact that (on the optimal half of the space) \(t_2 = t_1 + ||p_c - p_{c^\prime}|| \implies \) \(t_2 > t_1\) within \(L_p\). In such a case the global minimum lies at ``infinity" away from \(p_c\). When the global minimum is far away under these conditions, the optimization landscape will quickly plateau to zero and compactness suffers as the loss drops to zero and many of the embeddings move outwards without any coordination (e.g. Fig. \ref{fig:Optimization-Landscape-Convex}). Concave functions like in Fig. \ref{fig:Optimization-Landscape-Concave} suffer from the opposite problem and demonstrate the other extreme (\(\frac{df_1}{dt} > \frac{df_2}{dt}\)). In this case embeddings will anchor themselves toward their own proxy (the global minimum) and separability suffers. 

We now propose a simple yet effective example of a warping function which aims to strike a balance.
We select a point \(\alpha\) on \(L_p\) away from \(p_c\) but \(< \infty\) to serve as the global minimum, define \(f_2(t_2) = t_2\), and choose \(f_1\) so that the inequality of their derivatives changes directly at this point. 
\begin{equation}
\begin{aligned}
f_1(t_1) = \left\{
        \begin{array}{ll}
            k_{1}t_1 + \Delta & \quad 0 \leq t_1 < \alpha \\
            k_{2}t_1 + (1 - k_{2})\alpha & \quad \alpha \leq t_1
        \end{array}
    \right.\\
\end{aligned}
\label{eq:Warp-Function}
\end{equation}
where \(0 < k_1 < 1 < k_2\) are hyperparameters chosen to satisfy the above constraints. \(\Delta\) is added to counteract a negative side effect of the warp. 
Since \(k_1 < 1\), it artificially lowers (handicaps) the loss on the \([0,\alpha)\) interval by decreasing the distance between \(e\) and its ground-truth proxy \(p_c\). \(\Delta\) corrects this handicap by setting \(\Delta = t_1 - k_1t_1\) (with no back-propagated gradients). It can also be any value greater than this to help further bound the loss. The \((1 - k_{2})\alpha\) term is added for continuity. See Fig. \ref{fig:Optimization-Landscape-Warped} for an example warp. 

\begin{figure*}[t]
    \includegraphics[width=\textwidth, height=3.5cm]{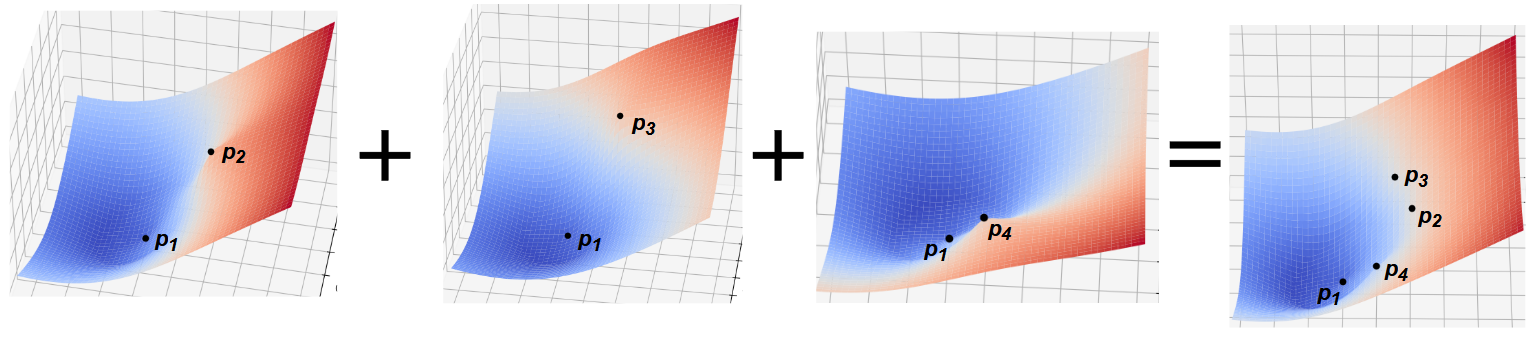}
    \caption{In the multi-class setting the binary class analysis can be repeated for each \((p_{y_i},p_j)\) proxy pair (\(y_i\) is the ground-truth). The optimization landscapes can then be "combined" to get the multi-class landscape.}
    \label{fig:Multi-Class-Landscape}
\end{figure*}

So far, we have assumed static proxies and focused solely on the forces acting on \(e\) (F1 and F3). However, there are still forces acting on \(p_c\) (F4) and \(p_{c^\prime}\) (F2).
Both aim to move their proxy closer to/further away from \(e\), respectively. Normally, these forces could threaten to subvert the forces acting on \(e\). However, the analysis so far is still valid since the Proposition/Lemma still applies regardless of the proxies' location. The warping then helps to mitigate this issue by realigning the forces on \(e\) with those on \(p_c\) and \(p_{c^\prime}\) so that they \emph{complement} one another (see Fig. \ref{fig:Aligned-Forces}). 

\subsection{Multi-Class Setting}
\label{subsec:Multi-Class-Setting}
We have also assumed a two-class setting. The multi-class case can simply be extended as follows: 
\begin{equation}
     CE = \log \Big(1 + \sum_{j \neq y_i}^{\mathcal{C}}\exp\big(f_1(||e_i - p_{y_i}||) - f_2(||e_i - p_j||)\big)\Big)    
     \label{eq:Multi-Class-Warp}
\end{equation}
For an embedding vector \(e_i\), the binary-class analysis can be repeated for each non-ground truth proxy (each \(f_1 - f_2\) term) for a total of \(\mathcal{C} - 1\) coupled minimums. The global minimum for \(e_i\) then simply becomes a combination of each one. 
As the 4 class example in Fig. \ref{fig:Multi-Class-Landscape} illustrates, the warped landscapes for all proxy pairs ``combine" together to form the multi-class landscape. This includes all the warped regions. It is possible the proxies won't be as conveniently arranged/placed as in the figure, but the inherent assumption in this work is that they \emph{learn} to reorganize themselves into the more optimal locations as training progresses. As such, the landscapes will start to combine in ways that complement each other and that still preserve a functioning warp even within the multi-class domain. 
Evidence for this can be found in the experiments and ablations in the following sections which were all performed within the multi-class setting and which clearly demonstrate the superiority of warped landscapes. 

Additionally, although it is certainly possible to assign distinct hyperparameter values for each class (potential future work), we would like to clarify that in this work we use the same hyperparameters uniformly across all classes. For example, each binary landscape is warped by the same \(k_1, k_2,\) and \(\alpha\) hyperparameters before being summed together in Eq. \ref{eq:Multi-Class-Warp}.  


\subsection{Interpretation of Proposed Loss}
\label{subsec:Loss-Interpretation}
The hyperparameters in Eq. \ref{eq:Warp-Function}/\ref{eq:Multi-Class-Warp} can be interpreted from the following perspective.
\(\frac{df_1}{dt} < \frac{df_2}{dt}\) boosts separability. \(\frac{df_1}{dt} > \frac{df_2}{dt}\) boosts compactness. Since \(f_2(t) = t, \frac{df_1}{dt} = k < 1\) boosts separability and \(k > 1\) compactness. 
As such, \(k_1 < 1\) determines the ``strength" of the outward forces. \(k_2 > 1\) determines the ``strength" of the inward forces. 
\(\alpha\) is the effective point of attraction (the minimum). It determines ``where" to apply the inward and outward forces respectively. 
\(\alpha = 0 \; (\infty)\) implies inward (outward) forces everywhere. 

\section{Experiments}

We demonstrate the effectiveness of our approach by evaluating it on various Image Retrieval benchmarks. 
We compare our method against recent metric learning loss functions and study the impact that various hyperparameters have on embedding space formation.

\begin{table*}[t]
\caption{Recall@K and NMI comparison on three Image-Retrieval datasets. \(\dagger\) denotes results on larger crop sizes. For the backbones: `I' abbreviates Inception with Batch Normalization \cite{ioffe2015batch} and `R' abbreviates ResNet50 \cite{he2016deep}. Superscripts denote embedding size. * denotes methods that used some additional components in their setups (compared to most other methods listed).}
\vspace{0.1in}
\centering
\begin{adjustbox}{width=1.0\textwidth}
\renewcommand{\arraystretch}{1.01}
\large
\begin{tabular}{ l | l | c  c  c  c | c  c  c  c | c  c  c  c }
     \hline
     \multicolumn{2}{c}{} & \multicolumn{4}{|c}{CUB-200-2011} & \multicolumn{4}{|c|}{CARS196} & \multicolumn{4}{c}{Stanford Online Products}\\
     \hline
     Method & BB & R@1 & R@2 & R@4 & NMI & R@1 & R@2 & R@4 & NMI & R@1 & R@10 & R@100 & R@1000\\
     \hline
     ProxyNCA \cite{movshovitz2017no} & I\textsuperscript{64} & 49.2 & 61.9 & 67.9 & 59.5 & 73.2 & 82.4 & 86.4 & 64.9 & 73.7 & - & - & -\\
     
     Multi-Sim \cite{wang2019multi} & I\textsuperscript{512} & 65.7 & 77.0 & 86.3 & - &  84.1 & 90.4 & 94.0 & - & 78.2 & 90.5 & 96.0 & 98.7\\

     N-Softmax \cite{zhai2018classification} & R\textsuperscript{512} & 61.3 & 73.9 & 83.5 & 69.7 &  84.2 & 90.4 & 94.4 & 74.0 & 78.2 & 90.6 & 96.2 & -\\

     SoftTriple \cite{qian2019softtriple} & I\textsuperscript{512} & 65.4 & 76.4 & 84.5 & 69.3 &  84.5 & 90.7 & 94.5 & 70.1 & 78.3 & 90.3 & 95.9 & -\\
     
     BlackBox \cite{rolinek2020optimizing} & R\textsuperscript{512} & 64.0 & 75.3 & 84.1 & - &  - & - & - & - & 78.6 & 90.5 & 96.0 & 98.7\\
     
     Circle Loss \cite{sun2020circle} & I\textsuperscript{512} & 66.7 & 77.4 & 86.2 & - &  83.4 & 89.8 & 94.1 & - & 78.3 & 90.5 & 96.1 & 98.6\\
     
     Proxy-Few \cite{zhu2020fewer} & I\textsuperscript{512} & 66.6 & 77.6 & 86.4 & 69.8 &  85.5 & 91.8 & 95.3 & 72.4 & 78.0 & 90.6 & 96.2 & -\\

     Proxy-Anchor \cite{kim2020proxy} & I\textsuperscript{512} & 68.4 & 79.2 & 86.8 & - &  86.1 & 91.7 & 95.0 & - & 79.1 & 90.8 & 96.2 & 98.7\\
     
     Proxy-Anchor \cite{kim2020proxy} & R\textsuperscript{512} & 69.7 & 80.0 & 87.0 & - &  87.9 & 93.0 & 96.1 & - & - & - & - & -\\
     
     \hline
     \multicolumn{7}{l}{\textbf{Recent Methods with a Validation Set:}}\\
     ProxyNCA++ \cite{teh2020proxynca++} & R\textsuperscript{512} & \underline{65.9} & \underline{77.3} & \underline{85.8} & \underline{71.1} &  \underline{84.1} & \underline{90.5} & \underline{94.4} & \underline{70.3} & 78.7 & 90.7 & 96.0 & \underline{98.5}\\
     
     RS@K \cite{patel2022recall} & R\textsuperscript{512} & - & - & - & - &  80.7 & 88.3 & 92.8 & - & \textbf{82.8} & \textbf{92.9} & \textbf{97.0} & \textbf{99.0}\\
     
     \rowcolor{lightgray}
     Warped-Softmax & R\textsuperscript{512} & \textbf{71.8} & \textbf{81.5} & \textbf{88.4} & \textbf{74.6} & \textbf{90.7} & \textbf{94.9} & \textbf{96.9} & \textbf{77.0} & \underline{81.6} & \underline{92.3} & \underline{96.7} & \textbf{99.0}\\
     
     \hline
     \multicolumn{7}{l}{\textbf{Recent Methods without a Validation Set:}}\\
     *CE \cite{boudiaf2020unifying} & R\textsuperscript{2048} & 69.2 & 79.2 & 86.9 & - & 89.3 & 93.9 & 96.6 & - & 81.1 & 91.7 & 96.3 & 98.8\\
     
     CBML \cite{kan2022contrastive} & R\textsuperscript{512} & 69.9 & 80.4 & 87.2 & 70.3 & 88.1 & 92.6 & 95.4 & 71.6 & 79.9 & 91.5 & 96.5 & 98.9\\

     IBC \cite{seidenschwarz2021learning} & R\textsuperscript{512} & 70.3 & 80.3 & 87.6 & 74.0 & 88.1 & 93.3 & 96.2 & 74.8 & 81.4 & 91.3 & 95.9 & -\\

     HIST \cite{lim2022hypergraph} & R\textsuperscript{512} & 71.4 & 81.1 & 88.1 & \underline{74.1} & 89.6 & 93.9 & 96.4 & 75.2 & 81.4 & 92.0 & \underline{96.7} & -\\

     *Contextual \cite{liao2023supervised} & R\textsuperscript{512} & \underline{71.9} & \underline{81.5} & \textbf{88.5} & - &  \underline{91.1} & \textbf{95.0} & \textbf{97.1} & - & \textbf{82.6} & \textbf{92.5} & \underline{96.7} & 98.8\\


     Softmax (Eq. \ref{eq:Default-Softmax-Inverted}) & R\textsuperscript{512} & 69.1 & 79.1 & 86.8 & 72.2 & 90.2 & 94.3 & 96.7 & \underline{75.9} & 79.5 & 91.0 & 96.2 & 98.8\\
     
     \rowcolor{lightgray}
     Warped-Softmax & R\textsuperscript{512} & \textbf{72.6} & \textbf{81.6} & \underline{88.4} & \textbf{74.6} & \textbf{91.2} & \underline{94.9} & \underline{96.9} & \textbf{77.2} & \underline{81.9} & \underline{92.3} & \textbf{96.8} & \textbf{99.0}\\
     \hline
     ROADMAP\({}^{\dagger}\)\cite{ramzi2021robust} & R\textsuperscript{512} & 68.5 & 78.7 & 86.6 & - &  - & - & - & - & \underline{83.1} & 92.7 & 96.3 & -\\
     
     OSAP\({}^{\dagger}\) \cite{yuan2023osap} & R\textsuperscript{512} & 70.5 & 80.3 & \underline{88.3} & - &  - & - & - & - & \textbf{84.4} & \underline{93.1} & \underline{97.3} & -\\

     *CCL\({}^{\dagger}\) \cite{cai2023center} & R\textsuperscript{512} & \textbf{73.5} & \underline{81.9} & 87.8 & - &  \underline{91.0} & \underline{94.5} & \underline{96.8} & - & \underline{83.1} & \textbf{93.3} & \textbf{97.4 }& -\\
     \rowcolor{lightgray}
     Warped-Softmax\({}^{\dagger}\)  & R\textsuperscript{512} & \underline{72.9} & \textbf{82.0} & \textbf{88.7} & \textbf{75.2} & \textbf{91.7} & \textbf{95.4} & \textbf{97.3} & \textbf{79.0} & 82.5 & 92.7 & 96.9 & \textbf{99.0}\\
     \bottomrule
\end{tabular}
\renewcommand{\arraystretch}{1}
\end{adjustbox}
\label{tab:Image-Retrieval-Results}
\end{table*}
\begin{table}[t]
\caption{Validation Set Results for our method in Table 1 but with standard deviations included. Other results contain similar deviations.}
\vspace{0.1in}
\centering
\begin{adjustbox}{width=1.0\columnwidth}
\large
\begin{tabular}{ l | c  c  c  c }
     \toprule
     Dataset & R@1 & R@2 & R@4 & NMI\\
     \hline
     CUB & $71.8 \pm 0.16$ & $81.5 \pm 0.17$ & $88.4 \pm 0.16$ & $74.6 \pm 0.12$\\
     CARS & $90.7 \pm 0.11$ & $94.9 \pm 0.06$ & $96.9 \pm 0.02$ & $77.0 \pm 0.32$\\
     \hline
     Dataset & R@1 & R@10 & R@100 & R@1000\\
     \hline
     SOP & $81.6 \pm 0.09$ & $92.3 \pm 0.05$ & $96.7 \pm 0.05$ & $99.0 \pm 0.02$\\
     \bottomrule
\end{tabular}
\end{adjustbox}
\label{tab:Image-Retrieval-Results-Stddev}
\end{table}

\subsection{Datasets}
\label{subsec:Datasets}
We evaluate our method on three standard image retrieval datasets. CUB-200-2011 \cite{welinder2010caltech} consists of 200 different classes of birds. The first 5,864 images belonging to the first 100 classes are used for training, and the last 5,924 images from the remaining classes for testing. 
Similarly, CARS-196 \cite{krause20133d} consists of
16,183 images divided into 196 classes of cars, the first 98 classes are used for training and the second 98 for testing. Stanford-Online-Products (SOP) \cite{oh2016deep} consists of 22,634 classes across 120,053 images. The dataset is split approximately half/half (11,318/11,316) into train/test partitions. 
The training set is further divided 50/50 into train/val splits in order to optimize hyperparameters. Afterwards, the full training set is used to train the model before evaluation on the official test split. 

\subsection{Implementation}
\label{subsec:Implementation}
\noindent\textbf{Setup:}
We build on top of ProxyNCA++ \cite{teh2020proxynca++} so we utilize a setup similar to theirs. This means using Layer Normalization \cite{ba2016layer}, class balanced sampling, temperature scaling, and a ResNet50 \cite{he2016deep} backbone with ImagNet \cite{deng2009imagenet} pre-trained weights. However, following precedent in \cite{kim2020proxy}, we use max + average pooling instead of just max pooling. We rerun ProxyNCA++ experiments to also use max + average pooling. As in \cite{movshovitz2017no, kim2020proxy, teh2020proxynca++}, we assign a single proxy per-class, and we initialize the proxies from a normal distribution. 
Images are scaled to 256x256 and then randomly cropped to 224x224 during training and center-cropped during testing for small sizes.
Experiments are also run on large (288/256) image/crop sizes. 
We further augment data with random horizontal flipping during training.

A few years ago \cite{musgrave2020metric} published a reality check of the field and made a number of criticisms of DML papers. One of their primary criticisms included the lack of validation sets for tuning hyperpaparameters. 
Unfortunately, most DML papers have continued to optimize hyperparameters directly on the test set without any validation splits.
As such, we categorize recent results by validation set status, and we report our own results in both categories to compare with all methods.  
Evaluations under Musgrave et. al.'s methodology can be found in Appendix D. 


\noindent\textbf{Training:}
We trained the model in two phases. 
This was to primarily prioritize separability before switching over to compactness later in the run.
For \(\alpha\) in particular, this meant using a relatively larger value at first and lowering it later.
We chose to perform a second hyperparameter search (with tighter bounds) for the later phase. Full details can be found with our code. 
All hyperparameters were obtained via a combination of grid + random search (for results using a Bayesian search see Appendix D). 
We adopt the Adam Optimizer \cite{kingma2014adam}. The final results are averaged across 5 total training runs.

\subsection{Results Comparison and Discussion}
\label{subsec:Results-Comparison}
There are many works in DML that differ in their respective strategies to advance the field. 
Not all are directly comparable to one another.
For example, some methods (e.g. \cite{ebrahimpour2022multi,gurbuz2023generalized}) focus on developing and enhancing the base architecture to improve the model. Other works (e.g. \cite{zhang2023calibration,saberi2023deep,roth2022non, mohan2020moving,zhang2023threshold,kim2023hier,kirchhof2022non}) focus on pioneering new regularization methods, and still others like \cite{sanakoyeu2019divide,zhao2021towards,venkataramanan2021takes,ko2021learning,barbany2024procsim,ren2024learning} focus on developing new and enhanced training strategies/techniques. For even comparison, in a work introducing a new loss function, we constrain our discussion to comparisons with other loss functions. We also omit comparisons to ensemble-based methods (e.g. \cite{zheng2021deep,wang2023deep,zheng2021deep2,zabihzadeh2024ensemble}) 
and to methods like \cite{yang2022hierarchical,li2023robust,ren2024towards,furusawamean} that can be applied on top of arbitrary loss functions. Although  we emphasize that our results are competitive with/superior to nearly all of these methods.
The results are in Tab. \ref{tab:Image-Retrieval-Results}. We omit the standard deviations in Tab. \ref{tab:Image-Retrieval-Results} for readability. But we include them in Tab. \ref{tab:Image-Retrieval-Results-Stddev} for complementary purposes (for validation data results). 

\begin{table*}[b]
    \caption{The impact of different Eq. \ref{eq:Multi-Class-Warp} functions on embedding space formation. \(t_1 - t_2\) is equivalent to Eq. \ref{eq:Default-Softmax-Inverted}. **.** denotes the last result preceding numerical instability.}
    \vspace{0.1in}
    \centering
    \begin{adjustbox}{width=0.85\textwidth, height=1.15\Height}
    \begin{tabular}{ l | c  c  c } 
         \hline
         \multicolumn{4}{c}{Function Comparisons}\\
         \hline
         \(f_1(t_1) - f_2(t_2)\) & R@1 & NMI & AvgDTP\\
         \hline

         Pretrained (ImageNet) & 51.8 & 60.8 & 13.40\\
    
         \(t_1 - t_2\) \;\; (Eq. \ref{eq:Default-Softmax-Inverted}) & 63.2 & 67.1 & 12.88\\
         
         \(t_1^2 - t_2^2\) & 57.2 & 64.9 & 14.67\\ 
         
         \(\sqrt{t_1} - \sqrt{t_2}\) & 61.6 & 68.1 & 6.48 \\
    
         \(t_1^{3/2} - t_2^{3/2}\) & 62.2 & 68.5 & 17.20\\ 
         
         **\(0.97t_1 - t_2\)** & 39.4 & 44.0 & 84.80\\

         **\(0.5t_1 - t_2\)** & 13.6 & 26.3 & 93.72\\
         
         \(1.1t_1 - t_2\) & 60.5 & 64.4 & 3.86\\
         
         \(2.0t_1 - t_2\) & 54.1 & 59.8 & 2.18\\
         \hline
    \end{tabular}
    \quad \quad
    \begin{tabular}{ c | c  c  c } 
         \hline
         \multicolumn{4}{c}{Eq. \ref{eq:Warp-Function} Warp Comparisons}\\
         \hline
         \(f_1(t_1) - t_2\) & R@1 & NMI & AvgDTP\\
         \hline
         $(7) \; \Biggl\{\begin{array} {r@{}l@{}} 
             & {} k_1 = 0.5\\ 
             & {} k_2 = 1.5\\
             &\alpha = 50.0
         \end{array}$ 
         & 18.6 & 28.7 & 50.05\\

         $(7) \; \Biggl\{\begin{array} {r@{}l@{}} 
             & {} k_1 = 0.9\\ 
             & {} k_2 = 1.1\\
             &\alpha = 25.0
         \end{array}$ 
         & 53.9 & 60.1 & 25.38\\
         
         $(7) \; \Biggl\{\begin{array} {r@{}l@{}} 
             & {} k_1 = 0.25\\ 
             & {} k_2 = 2.25\\
             &\alpha = 7.75
         \end{array}$ 
         
         & 64.6 & 68.5 & 7.76\\
         \hline
    \end{tabular}
    \end{adjustbox}
    \label{tab:Function-Comparison}
\end{table*}
\begin{figure}[t]
    \centering
    \includegraphics[width=0.8\columnwidth, height=5.0cm]{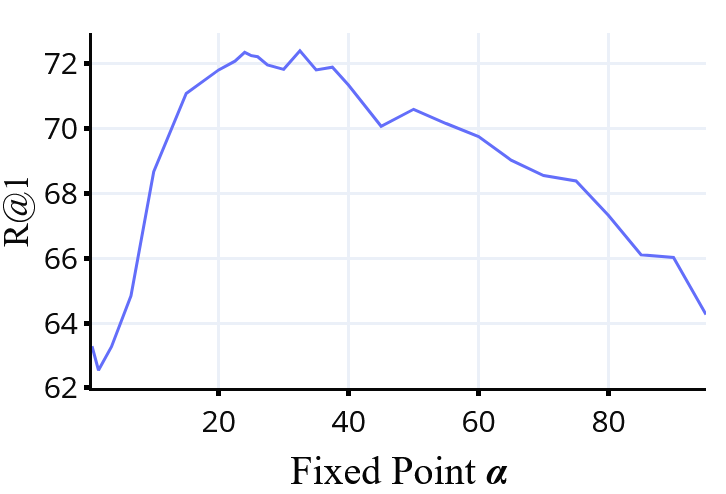}
    \caption{Impact of the warp parameter \(\alpha\) on R@1 (CUB-200).}
    \label{fig:Hyperparameters-Alpha}
\end{figure}

In accordance with most metric learning papers, we make use of the standard Recall@K and NMI metrics. Results are categorized by validation set use and crop size.
Our results are the overall best on CUB and a mixture of competitive/superior for Cars and SOP depending on the metric and category.  
Some methods (* in Tab. \ref{tab:Image-Retrieval-Results}) use extra components in their setups (e.g. random photometric jittering of data, drop out, extra linear/batchnorm layers on the output) compared to most others listed. 
Our results are still competitive with/superior to these methods. 
We also significantly outperform ProxyNCA++, the closest related method to ours (minus warped landscapes and the base euclidean metric), by large margins across all datasets. 
Additionally, as an ablation to test out the overall effectiveness of warping, we compare our method to a default Eq. \ref{eq:Default-Softmax-Inverted} (non-Warped) euclidean softmax on the smaller sizes. The warped softmax outperforms the default for all metrics on all datasets.



\subsection{Hyperparameter Impact}
\label{subsec:Hyperparameter-Impact}

\begin{figure}[t]
    \centering
    \includegraphics[width=\columnwidth, height=5.0cm]{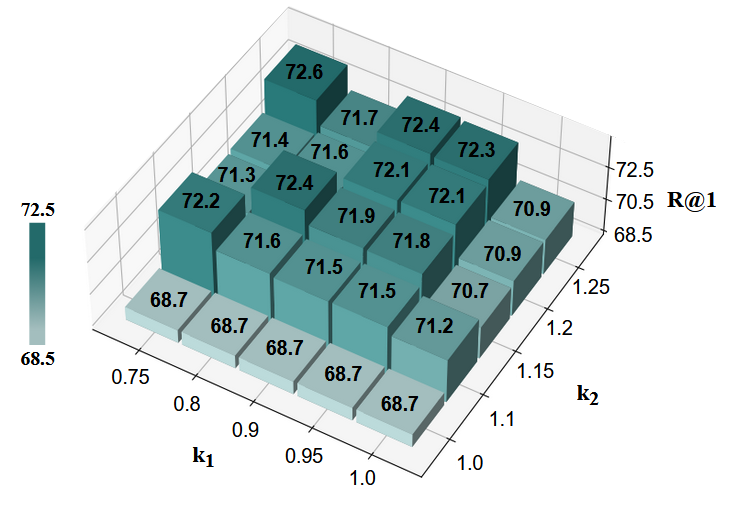}
    \caption{Impact of the warp parameters \(k_1 \text{ and } k_2,\)  on R@1 for (CUB-200).}
    \label{fig:Hyperparameters-k12}
\end{figure}
\begin{figure}[t]
    \centering
    \includegraphics[width=0.90\columnwidth, height=4.5cm]{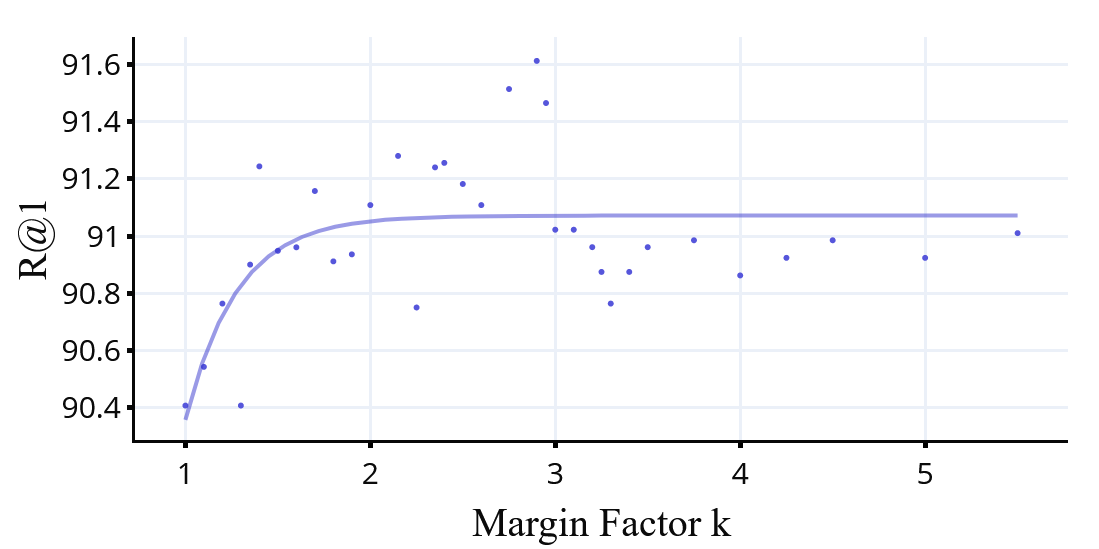}
    \caption{Varying \(\Delta\) in Eq. \ref{eq:Warp-Function} on Cars196 with \(k\Delta\) (\(k \geq 1\)).}
    \label{fig:Hyperparameters-Delta}
\end{figure}

Using CUB, we investigate the impact of \(k_1\), \(k_2\), and \(\alpha\) on the final accuracy. For this, we fix the other hyperparameters (e.g. learning rates, temperature) to the same values used to produce the results in Tab. \ref{tab:Image-Retrieval-Results} and run two experiments. For the first, we fix \(k_{1,2}\) like the others, and incrementally vary \(\alpha\) from 0 to 95.
The results are shown in Fig. \ref{fig:Hyperparameters-Alpha}.
They suggest that the accuracy increases, begins to stabilize for a period and then drops off. 
For the second experiment, we fix \(\alpha\) and vary \(k_{1,2}\), and report the changes in accuracy.
These results are illustrated in Fig. \ref{fig:Hyperparameters-k12}. 
Here, accuracy begins to roughly stabilize at higher values when neither \(k_1\) nor \(k_2\) are 1.0. This makes sense because the boost for the outward (inward) forces is effectively turned off when \(k_1\) (\(k_2\)) is 1.0. 
Otherwise the values begin to matter somewhat less.


As explained in Section \ref{subsec:Function-Warping}, the primary purpose of \(\Delta\) in Eq. \ref{eq:Warp-Function} is to counteract the handicap induced by \(k_1\). However, \(\Delta\) be can be increased even more to further bound the loss (thus serving more like a traditional margin). We study the impact of doing this by multiplying it by a constant \(k \geq 1\) and by varying this constant from 1 to 5.5. The results are shown in Fig. \ref{fig:Hyperparameters-Delta}. According to these results, there is an initial trend where increasing \(\Delta\) improves the results. However, past a certain point accuracy begins to drop and then stabilize and plateau.

\subsection{Function Ablation}
\label{subsec:Function-Ablation}
\begin{figure*}[t]
    \centering
    \subfloat[]{
        \includegraphics[width=0.45\textwidth, height=6.0cm]{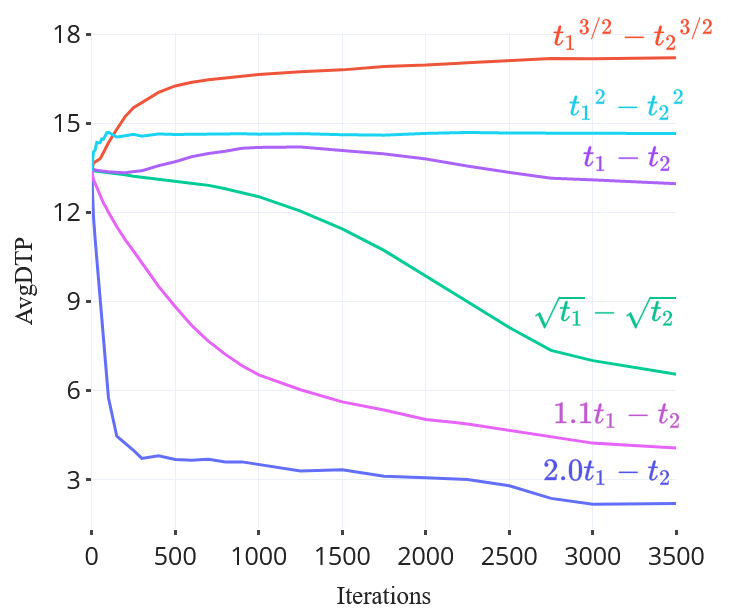}
    }
    \hfill
    \subfloat[]{
        \includegraphics[width=0.5\textwidth, height=6.0cm]{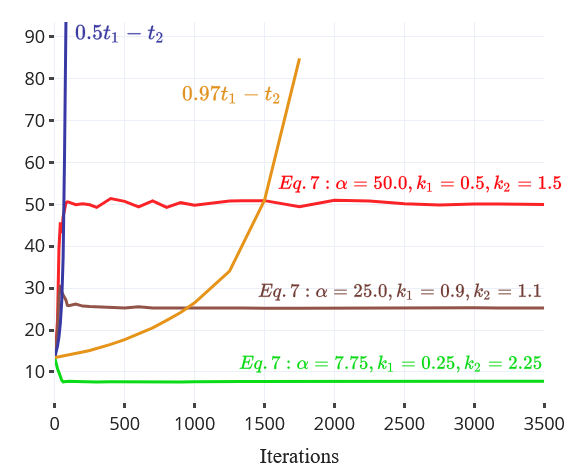}
    }
    \caption{The behavior of various functions over the course of a training run under CUB. Corresponding functions can be found in Table \ref{tab:Function-Comparison}.}
    \label{fig:Function-Behaviour}
\end{figure*}

Additionally, to gain insight over the way different functions impact embedding space formation, we track the relative positioning between embeddings and ground-truth proxies over the course of a run on CUB.
\begin{equation}
    AvgDTP = \frac{1}{\mathcal{C}} \sum_{c = 1}^{\mathcal{C}} \frac{1}{|P_c|} \sum_{e \in P_c} ||e - p_c||
\end{equation}
DTP abbreviates \textit{Distance To Proxy}, and \(P_c = \{e_i | y_i = c\}\).

Tab. \ref{tab:Function-Comparison} displays the results various functions substituted into Eq. \ref{eq:Multi-Class-Warp}, and Fig. \ref{fig:Function-Behaviour} plots their DTP value over a run. The traditional metrics (R@1 and NMI) are calculated off of test data while AvgDTP is calculated using training samples. We do not apply data augmentation in order to better observe the direct behavior induced by each function. And we initialize the last fully connected layer slightly different so that embeddings start off closer to their proxies. 
The functions differ in which force they prioritize during training. Some (e.g. the convex ones) prioritize separability. Others (e.g. \(\sqrt{t_1} - \sqrt{t_2}\)) prioritize compactness. Fig. \ref{fig:Function-Behaviour} illustrates these differences. Functions that prioritize compactness over separability are correlated with a small DTP at the end of the run. The examples here contain minimums directly on their proxies. Consequently, the model focuses too hard on pushing same-class embeddings together towards their proxy and the final results suffer. 
The functions that prioritize separability over compactness suffer from the opposite problem and correlate with a higher DTP.

The most optimal vanilla choice is the default Eq. \ref{eq:Default-Softmax-Inverted} softmax (\(t_1 - t_2\)).
As shown in Fig. \ref{fig:Function-Behaviour}, it contains a relatively steady DTP, although it fluctuates slightly as embeddings move to their proper locations within the space.
With proper settings, a Softmax Warp maintains a constant balance between the forces of separability and compactness and produces more optimal results (note the correlation between \(\alpha\) and DTP). Also note the ``half-warp" cases in Tab. \ref{tab:Function-Comparison}. For ``\(0.97t_1 - t_2\)" and ``\(0.5t_1 - t_2\)" there are outward forces everywhere since \(\frac{df_1}{dt} < \frac{df_2}{dt}\). 
Additionally, the \(k_1\) term here boosts these forces significantly. 
As such, the embeddings fly off to infinity until numerical instability (note the high DTP values). The opposite case is true for 
the ``\(1.1t_1 - t_2\)" and ``\(2.0t_1 - t_2\)" cases.

\subsection{Noise-Robustness}
\label{subsec:Noise-Robustness}
Given that our methodology is constructed within a Euclidean setting, we ran the following experiments using the imagaug \cite{imgaug} library to investigate how robust the unbounded space is compared to loss functions that operate within a bounded cosine/normalized space. 
We compared the performance of our Warped Softmax (Eq. \ref{eq:Warp-Function}/\ref{eq:Multi-Class-Warp}) against that of the Vanilla Softmax (Eq. \ref{eq:Default-Softmax-Inverted}), the l2-normalized method most closely related to ours (ProxyNCA++ \cite{teh2020proxynca++}), and a more recent method (HIST \cite{lim2022hypergraph}). The results are in Table \ref{tab:Noise-Corruption-Test}. We used the default settings for most of the corruptions. Please see \cite{imgaug} for more information on each type of corruption. Our loss still outperforms ProxyNCA++ and the Vanilla Softmax on the Noisy/Corrupted Data and outperformed HIST on all but Cutout and Jigsaw. 

\begin{table}[t]
\caption{R@1 values for different losses on Noisy/Corrupted Data.}
\vspace{0.1in}
\centering
\begin{adjustbox}{width=1.0\columnwidth}
\renewcommand{\arraystretch}{1.075}
\Huge
\begin{tabular}{ l  c  c  c  c}
     \hline
     Corruption & HIST & ProxyNCA++ & Regular Softmax & Warped Softmax\\

     \hline
     Uncorrupted & \underline{71.8} & 66.3 & 69.4 & \textbf{72.5} \\
     
     \hline
     \multicolumn{3}{l}{\textbf{Additive Noise:}}\\
     Gaussian & \underline{57.2} & 51.5 & 56.7 & \textbf{57.7}\\
     
     Uniform  & \underline{68.7} & 63.0 & 66.9 & \textbf{69.1} \\
     
     Salt and Pepper & \underline{39.5} & 33.8 & 39.4 & \textbf{41.0}\\

     \hline
     \multicolumn{3}{l}{\textbf{Dropping Pixels:}}\\
     Dropout & \underline{27.5} & 22.5 & 27.3 & \textbf{28.9} \\
     
     Cutout & \textbf{64.5} & 54.5 & 59.5 & \underline{61.6} \\

     \hline
     \multicolumn{3}{l}{\textbf{Geometric Distortions:}}\\
     Affine Shear & \underline{67.5} & 63.4 & 67.4 & \textbf{69.7} \\

     Jigsaw & \textbf{56.4} & 45.7 & 51.2 & \underline{55.6} \\

     \hline
     \multicolumn{3}{l}{\textbf{Color Distortions:}}\\
     Add Hue and Saturation & \underline{50.2} & 44.9 & 49.0 & \textbf{51.1} \\

     \hline
     \multicolumn{3}{l}{\textbf{Blur:}}\\
     
     Defocus Blur & \underline{46.2} & 44.4 & 46.1 & \textbf{47.6} \\
     
     \bottomrule
\end{tabular}
\renewcommand{\arraystretch}{1}
\end{adjustbox}
\label{tab:Noise-Corruption-Test}
\end{table}

\subsection{Toy Illustration}
\label{subsec:Toy-Illustration}
\begin{figure}[t]
    \centering
    \subfloat[Normal \label{fig:Toy-Example-Normal}]{
        \includegraphics[width=0.45\columnwidth, height=3.25cm]{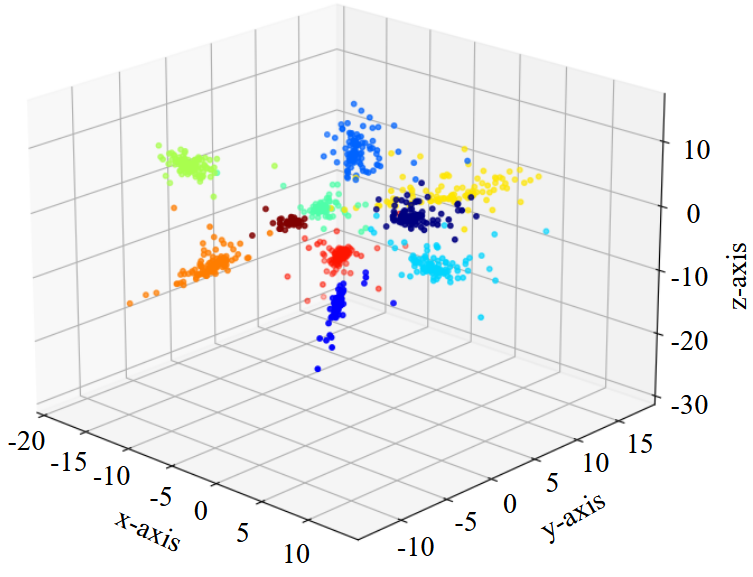}
    }
    \hspace{0.25cm}
    \subfloat[Warped \label{fig:Toy-Example-Warped}]{
        \includegraphics[width=0.45\columnwidth, height=3.25cm]{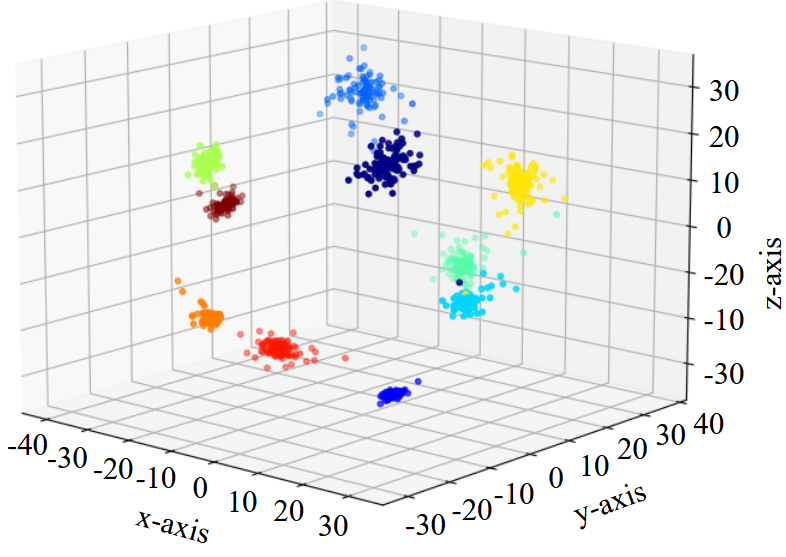}
    }
    \caption{Toy Example illustrating a normal softmax (Eq. \ref{eq:Default-Softmax-Inverted}) vs. our method (Eq. \ref{eq:Warp-Function} \& \ref{eq:Multi-Class-Warp}) on embedding formation. Embeddings are trained with a modified LeNet model on FashionMNIST. The warped embedding leads to better-separated clusters while preserving compactness.}
    \label{fig:Toy-Example-Official}
\end{figure}

Additionally, to further illustrate the effect of our method on embedding formation, we train example overfits on a 5-layer CNN using FashionMNIST and our proposed loss vs a default softmax and compare the difference. 
Fig. \ref{fig:Toy-Example-Official} shows the result. 
Compared to a normal softmax, our warped method works to better separate the clusters and preserve compactness. 

\subsection{Face Recognition}

The field of face recognition has been largely dominated by cosine-based losses \cite{deng2019arcface,wang2018cosface,liu2017sphereface,meng2021magface}.
A euclidean-based loss could open up another avenue of exploration for face recognition loss functions. As such, to potentially stimulate further interest we study the performance of our method on some standard face recognition datasets.

\subsubsection{Setup}
Following in the footsteps of \cite{deng2019arcface} and \cite{meng2021magface} we employ MS1MV-2 \cite{deng2019arcface} (about 5.8M images and 85k identities) as the training set. For evalution we use the standard benchmarks LFW \cite{huang2008labeled}, CFP-FP \cite{sengupta2016frontal}, AgeDB-30 \cite{moschoglou2017agedb}, CALFW \cite{zheng2017cross}, and CPLFW \cite{zheng2018cross} and report the verification accuracy.
We re-implement our method along with some classic face recognition loss functions within the codebase of \cite{deng2019arcface} and use the recommended hyperparameters for each. 
A ResNet100 is universally employed as the backbone and each model is trained on 4-8 (same batch size/iterations) A100 GPUs for 25 epochs using stochastic gradient descent.
Full details can be found in our code. 
\begin{table}[t]
    \caption{Results on Standard Benchmarks}
    \vspace{0.1in}
    \centering
    \begin{adjustbox}{width=1.0\columnwidth}
    \renewcommand{\arraystretch}{1.40}
    \large
    \begin{tabular}{ l | c  c  c  c  c } 
     \hline
     Method & LFW & CFP-FP & AgeDB-30 & CALFW & CPLFW\\
     \hline
     Softmax (cos) & 99.33 & 93.49 & 94.25 & 93.78 & 88.00\\
     
     SphereFace \cite{liu2017sphereface} & 99.80 & \underline{98.51} & 98.10 & \textbf{96.23} & \underline{93.18}\\

     Arcface \cite{deng2019arcface} & 99.82 & 98.49 & 98.18 & 96.10 & 92.95\\

     CosFace \cite{wang2018cosface} & 99.78 & \underline{98.51} & \textbf{98.32} & 96.17 & 92.93\\
     
     MagFace \cite{meng2021magface} & \underline{99.83} & 98.26 & \underline{98.23} & \underline{96.18} & 92.92\\
     
     Softmax (Eq. \ref{eq:Default-Softmax-Inverted}) & 99.55 & 97.10 & 97.38 & 94.72 & 91.17\\

     \rowcolor{lightgray}
     Warped-Softmax &\textbf{99.85} & \textbf{98.66} & 97.63 & 95.82 & \textbf{93.25}\\
     \hline
    \end{tabular}
    \renewcommand{\arraystretch}{1}
    \label{tab:Face-Recognition-Standard}
    \end{adjustbox}
\end{table}

\subsubsection{Results}
The results are displayed in Tab. \ref{tab:Face-Recognition-Standard}. For these standard benchmarks, we achieve competitive, if not superior, performance compared with some recent cosine-based loss functions. And we vastly improve over the performance of the default Eq. \ref{eq:Default-Softmax-Inverted} softmax which fails to be competitive. However, we acknowledge more work is necessary to thoroughly evaluate and fine-tune this approach for practical application (which we leave to the future).


\section{Conclusion}
In this work, we have introduced some basic theory describing the behavior of the coupled forces at work within a euclidean softmax embedding. We have explained how to formulate a potential new set of euclidean warp-based loss functions and provided a simple example. With this simple example, we have demonstrated the success and competitiveness of this approach on various DML benchmarks. 
However, the room for improvement is still large. 
Future work can range from investigating more advanced warping functions to more generalized base metrics as foundations. \\

\noindent\textbf{Acknowledgements:}
This research made use of the resources of the High Performance Computing Center at Idaho National Laboratory, which is supported by the Office of Nuclear Energy of the U.S. Department of Energy and the Nuclear Science User Facilities under Contract No. DE-AC07-05ID14517. We also thank Robert (Bob) Tarwater at INL for his continued assistance and backing of this work.

{
\small
\nocite{demoor2024realignedsoftmaxwarpingdeep}
\bibliographystyle{unsrt}
\bibliography{egbib}

\begin{thebibliography}{10}

\bibitem{movshovitz2017no}
Yair Movshovitz-Attias, Alexander Toshev, Thomas~K Leung, Sergey Ioffe, and Saurabh Singh.
\newblock No fuss distance metric learning using proxies.
\newblock In {\em Proceedings of the IEEE International Conference on Computer Vision}, pages 360--368, 2017.

\bibitem{teh2020proxynca++}
Eu~Wern Teh, Terrance DeVries, and Graham~W Taylor.
\newblock Proxynca++: Revisiting and revitalizing proxy neighborhood component analysis.
\newblock In {\em European Conference on Computer Vision}, pages 448--464. Springer, 2020.

\bibitem{kim2020proxy}
Sungyeon Kim, Dongwon Kim, Minsu Cho, and Suha Kwak.
\newblock Proxy anchor loss for deep metric learning.
\newblock In {\em Proceedings of the IEEE/CVF Conference on Computer Vision and Pattern Recognition}, pages 3238--3247, 2020.

\bibitem{sun2020circle}
Yifan Sun, Changmao Cheng, Yuhan Zhang, Chi Zhang, Liang Zheng, Zhongdao Wang, and Yichen Wei.
\newblock Circle loss: A unified perspective of pair similarity optimization.
\newblock In {\em Proceedings of the IEEE/CVF Conference on Computer Vision and Pattern Recognition}, pages 6398--6407, 2020.

\bibitem{deng2019arcface}
Jiankang Deng, Jia Guo, Niannan Xue, and Stefanos Zafeiriou.
\newblock Arcface: Additive angular margin loss for deep face recognition.
\newblock In {\em Proceedings of the IEEE/CVF conference on computer vision and pattern recognition}, pages 4690--4699, 2019.

\bibitem{wang2018cosface}
Hao Wang, Yitong Wang, Zheng Zhou, Xing Ji, Dihong Gong, Jingchao Zhou, Zhifeng Li, and Wei Liu.
\newblock Cosface: Large margin cosine loss for deep face recognition.
\newblock In {\em Proceedings of the IEEE conference on computer vision and pattern recognition}, pages 5265--5274, 2018.

\bibitem{liao2022graph}
Shengcai Liao and Ling Shao.
\newblock Graph sampling based deep metric learning for generalizable person re-identification.
\newblock In {\em Proceedings of the IEEE/CVF Conference on Computer Vision and Pattern Recognition}, pages 7359--7368, 2022.

\bibitem{chen2017beyond}
Weihua Chen, Xiaotang Chen, Jianguo Zhang, and Kaiqi Huang.
\newblock Beyond triplet loss: a deep quadruplet network for person re-identification.
\newblock In {\em Proceedings of the IEEE conference on computer vision and pattern recognition}, pages 403--412, 2017.

\bibitem{fathi2017semantic}
Alireza Fathi, Zbigniew Wojna, Vivek Rathod, Peng Wang, Hyun~Oh Song, Sergio Guadarrama, and Kevin~P Murphy.
\newblock Semantic instance segmentation via deep metric learning.
\newblock {\em arXiv preprint arXiv:1703.10277}, 2017.

\bibitem{neven2019instance}
Davy Neven, Bert~De Brabandere, Marc Proesmans, and Luc~Van Gool.
\newblock Instance segmentation by jointly optimizing spatial embeddings and clustering bandwidth.
\newblock In {\em Proceedings of the IEEE/CVF Conference on Computer Vision and Pattern Recognition}, pages 8837--8845, 2019.

\bibitem{ebrahimpour2022multi}
Mohammad~K Ebrahimpour, Gang Qian, and Allison Beach.
\newblock Multi-head deep metric learning using global and local representations.
\newblock In {\em Proceedings of the IEEE/CVF Winter Conference on Applications of Computer Vision}, pages 3031--3040, 2022.

\bibitem{elezi2020group}
Ismail Elezi, Sebastiano Vascon, Alessandro Torcinovich, Marcello Pelillo, and Laura Leal-Taix{\'e}.
\newblock The group loss for deep metric learning.
\newblock In {\em European Conference on Computer Vision}, pages 277--294. Springer, 2020.

\bibitem{sanakoyeu2019divide}
Artsiom Sanakoyeu, Vadim Tschernezki, Uta Buchler, and Bjorn Ommer.
\newblock Divide and conquer the embedding space for metric learning.
\newblock In {\em Proceedings of the IEEE/CVF Conference on Computer Vision and Pattern Recognition}, pages 471--480, 2019.

\bibitem{wu2017sampling}
Chao-Yuan Wu, R~Manmatha, Alexander~J Smola, and Philipp Krahenbuhl.
\newblock Sampling matters in deep embedding learning.
\newblock In {\em Proceedings of the IEEE international conference on computer vision}, pages 2840--2848, 2017.

\bibitem{sohn2016improved}
Kihyuk Sohn.
\newblock Improved deep metric learning with multi-class n-pair loss objective.
\newblock {\em Advances in neural information processing systems}, 29, 2016.

\bibitem{oh2016deep}
Hyun Oh~Song, Yu~Xiang, Stefanie Jegelka, and Silvio Savarese.
\newblock Deep metric learning via lifted structured feature embedding.
\newblock In {\em Proceedings of the IEEE conference on computer vision and pattern recognition}, pages 4004--4012, 2016.

\bibitem{hadsell2006dimensionality}
Raia Hadsell, Sumit Chopra, and Yann LeCun.
\newblock Dimensionality reduction by learning an invariant mapping.
\newblock In {\em 2006 IEEE Computer Society Conference on Computer Vision and Pattern Recognition (CVPR'06)}, volume~2, pages 1735--1742. IEEE, 2006.

\bibitem{schroff2015facenet}
Florian Schroff, Dmitry Kalenichenko, and James Philbin.
\newblock Facenet: A unified embedding for face recognition and clustering.
\newblock In {\em Proceedings of the IEEE conference on computer vision and pattern recognition}, pages 815--823, 2015.

\bibitem{harwood2017smart}
Ben Harwood, Vijay Kumar~BG, Gustavo Carneiro, Ian Reid, and Tom Drummond.
\newblock Smart mining for deep metric learning.
\newblock In {\em Proceedings of the IEEE International Conference on Computer Vision}, pages 2821--2829, 2017.

\bibitem{he2020softmax}
Lanqing He, Zhongdao Wang, Yali Li, and Shengjin Wang.
\newblock Softmax dissection: Towards understanding intra-and inter-class objective for embedding learning.
\newblock In {\em Proceedings of the AAAI Conference on Artificial Intelligence}, volume~34, pages 10957--10964, 2020.

\bibitem{zhai2018classification}
Andrew Zhai and Hao-Yu Wu.
\newblock Classification is a strong baseline for deep metric learning.
\newblock {\em arXiv preprint arXiv:1811.12649}, 2018.

\bibitem{qian2019softtriple}
Qi~Qian, Lei Shang, Baigui Sun, Juhua Hu, Hao Li, and Rong Jin.
\newblock Softtriple loss: Deep metric learning without triplet sampling.
\newblock In {\em Proceedings of the IEEE/CVF International Conference on Computer Vision}, pages 6450--6458, 2019.

\bibitem{boudiaf2020unifying}
Malik Boudiaf, J{\'e}r{\^o}me Rony, Imtiaz~Masud Ziko, Eric Granger, Marco Pedersoli, Pablo Piantanida, and Ismail~Ben Ayed.
\newblock A unifying mutual information view of metric learning: cross-entropy vs. pairwise losses.
\newblock In {\em European conference on computer vision}, pages 548--564. Springer, 2020.

\bibitem{yang2023stop}
Lu~Yang, Peng Wang, and Yanning Zhang.
\newblock Stop-gradient softmax loss for deep metric learning.
\newblock In {\em Proceedings of the AAAI Conference on Artificial Intelligence}, volume~37, pages 3164--3172, 2023.

\bibitem{liu2016large}
Weiyang Liu, Yandong Wen, Zhiding Yu, and Meng Yang.
\newblock Large-margin softmax loss for convolutional neural networks.
\newblock {\em arXiv preprint arXiv:1612.02295}, 2016.

\bibitem{ranjan2017l2}
Rajeev Ranjan, Carlos~D Castillo, and Rama Chellappa.
\newblock L2-constrained softmax loss for discriminative face verification.
\newblock {\em arXiv preprint arXiv:1703.09507}, 2017.

\bibitem{welinder2010caltech}
Peter Welinder, Steve Branson, Takeshi Mita, Catherine Wah, Florian Schroff, Serge Belongie, and Pietro Perona.
\newblock Caltech-ucsd birds 200.
\newblock 2010.

\bibitem{krause20133d}
Jonathan Krause, Michael Stark, Jia Deng, and Li~Fei-Fei.
\newblock 3d object representations for fine-grained categorization.
\newblock In {\em Proceedings of the IEEE international conference on computer vision workshops}, pages 554--561, 2013.

\bibitem{weinberger2005distance}
Kilian~Q Weinberger, John Blitzer, and Lawrence Saul.
\newblock Distance metric learning for large margin nearest neighbor classification.
\newblock {\em Advances in neural information processing systems}, 18, 2005.

\bibitem{wang2019multi}
Xun Wang, Xintong Han, Weilin Huang, Dengke Dong, and Matthew~R Scott.
\newblock Multi-similarity loss with general pair weighting for deep metric learning.
\newblock In {\em Proceedings of the IEEE/CVF Conference on Computer Vision and Pattern Recognition}, pages 5022--5030, 2019.

\bibitem{kan2022contrastive}
Shichao Kan, Zhiquan He, Yigang Cen, Yang Li, Vladimir Mladenovic, and Zhihai He.
\newblock Contrastive bayesian analysis for deep metric learning.
\newblock {\em IEEE Transactions on Pattern Analysis and Machine Intelligence}, 2022.

\bibitem{patel2022recall}
Yash Patel, Giorgos Tolias, and Ji{\v{r}}{\'\i} Matas.
\newblock Recall@ k surrogate loss with large batches and similarity mixup.
\newblock In {\em Proceedings of the IEEE/CVF Conference on Computer Vision and Pattern Recognition}, pages 7502--7511, 2022.

\bibitem{revaud2019learning}
Jerome Revaud, Jon Almaz{\'a}n, Rafael~S Rezende, and Cesar Roberto~de Souza.
\newblock Learning with average precision: Training image retrieval with a listwise loss.
\newblock In {\em Proceedings of the IEEE/CVF International Conference on Computer Vision}, pages 5107--5116, 2019.

\bibitem{cakir2019deep}
Fatih Cakir, Kun He, Xide Xia, Brian Kulis, and Stan Sclaroff.
\newblock Deep metric learning to rank.
\newblock In {\em Proceedings of the IEEE/CVF conference on computer vision and pattern recognition}, pages 1861--1870, 2019.

\bibitem{ramzi2021robust}
Elias Ramzi, Nicolas Thome, Cl{\'e}ment Rambour, Nicolas Audebert, and Xavier Bitot.
\newblock Robust and decomposable average precision for image retrieval.
\newblock {\em Advances in Neural Information Processing Systems}, 34:23569--23581, 2021.

\bibitem{yuan2023osap}
Xin Yuan, Xin Xu, Xiao Wang, Kai Zhang, Liang Liao, Zheng Wang, and Chia-Wen Lin.
\newblock Osap-loss: Efficient optimization of average precision via involving samples after positive ones towards remote sensing image retrieval.
\newblock {\em CAAI Transactions on Intelligence Technology}, 2023.

\bibitem{brown2020smooth}
Andrew Brown, Weidi Xie, Vicky Kalogeiton, and Andrew Zisserman.
\newblock Smooth-ap: Smoothing the path towards large-scale image retrieval.
\newblock In {\em European Conference on Computer Vision}, pages 677--694. Springer, 2020.

\bibitem{rolinek2020optimizing}
Michal Rol{\'\i}nek, V{\'\i}t Musil, Anselm Paulus, Marin Vlastelica, Claudio Michaelis, and Georg Martius.
\newblock Optimizing rank-based metrics with blackbox differentiation.
\newblock In {\em Proceedings of the IEEE/CVF Conference on Computer Vision and Pattern Recognition}, pages 7620--7630, 2020.

\bibitem{liao2023supervised}
Christopher Liao, Theodoros Tsiligkaridis, and Brian Kulis.
\newblock Supervised metric learning to rank for retrieval via contextual similarity optimization.
\newblock 2023.

\bibitem{zhu2020fewer}
Yuehua Zhu, Muli Yang, Cheng Deng, and Wei Liu.
\newblock Fewer is more: A deep graph metric learning perspective using fewer proxies.
\newblock {\em Advances in Neural Information Processing Systems}, 33:17792--17803, 2020.

\bibitem{yang2022hierarchical}
Zhibo Yang, Muhammet Bastan, Xinliang Zhu, Douglas Gray, and Dimitris Samaras.
\newblock Hierarchical proxy-based loss for deep metric learning.
\newblock In {\em Proceedings of the IEEE/CVF Winter Conference on Applications of Computer Vision}, pages 1859--1868, 2022.

\bibitem{seidenschwarz2021learning}
Jenny~Denise Seidenschwarz, Ismail Elezi, and Laura Leal-Taix{\'e}.
\newblock Learning intra-batch connections for deep metric learning.
\newblock In {\em International Conference on Machine Learning}, pages 9410--9421. PMLR, 2021.

\bibitem{lim2022hypergraph}
Jongin Lim, Sangdoo Yun, Seulki Park, and Jin~Young Choi.
\newblock Hypergraph-induced semantic tuplet loss for deep metric learning.
\newblock In {\em Proceedings of the IEEE/CVF Conference on Computer Vision and Pattern Recognition}, pages 212--222, 2022.

\bibitem{meng2021magface}
Qiang Meng, Shichao Zhao, Zhida Huang, and Feng Zhou.
\newblock Magface: A universal representation for face recognition and quality assessment.
\newblock In {\em Proceedings of the IEEE/CVF Conference on Computer Vision and Pattern Recognition}, pages 14225--14234, 2021.

\bibitem{ioffe2015batch}
Sergey Ioffe and Christian Szegedy.
\newblock Batch normalization: Accelerating deep network training by reducing internal covariate shift.
\newblock In {\em International conference on machine learning}, pages 448--456. PMLR, 2015.

\bibitem{he2016deep}
Kaiming He, Xiangyu Zhang, Shaoqing Ren, and Jian Sun.
\newblock Deep residual learning for image recognition.
\newblock In {\em Proceedings of the IEEE conference on computer vision and pattern recognition}, pages 770--778, 2016.

\bibitem{cai2023center}
Bolun Cai, Pengfei Xiong, and Shangxuan Tian.
\newblock Center contrastive loss for metric learning.
\newblock {\em arXiv preprint arXiv:2308.00458}, 2023.

\bibitem{ba2016layer}
Jimmy~Lei Ba, Jamie~Ryan Kiros, and Geoffrey~E Hinton.
\newblock Layer normalization.
\newblock {\em arXiv preprint arXiv:1607.06450}, 2016.

\bibitem{deng2009imagenet}
Jia Deng, Wei Dong, Richard Socher, Li-Jia Li, Kai Li, and Li~Fei-Fei.
\newblock Imagenet: A large-scale hierarchical image database.
\newblock In {\em 2009 IEEE conference on computer vision and pattern recognition}, pages 248--255. Ieee, 2009.

\bibitem{musgrave2020metric}
Kevin Musgrave, Serge Belongie, and Ser-Nam Lim.
\newblock A metric learning reality check.
\newblock In {\em European Conference on Computer Vision}, pages 681--699. Springer, 2020.

\bibitem{kingma2014adam}
Diederik~P Kingma and Jimmy Ba.
\newblock Adam: A method for stochastic optimization.
\newblock {\em arXiv preprint arXiv:1412.6980}, 2014.

\bibitem{gurbuz2023generalized}
Yeti~Z G{\"u}rb{\"u}z, Ozan Sener, and A~Aydin Alatan.
\newblock Generalized sum pooling for metric learning.
\newblock In {\em Proceedings of the IEEE/CVF International Conference on Computer Vision}, pages 5462--5473, 2023.

\bibitem{zhang2023calibration}
Qin Zhang, Linghan Xu, Qingming Tang, Jun Fang, Ying~Nian Wu, Joe Tighe, and Yifan Xing.
\newblock Calibration-aware margin loss: Pushing the accuracy-calibration consistency pareto frontier for deep metric learning.
\newblock {\em arXiv preprint arXiv:2307.04047}, 2023.

\bibitem{saberi2023deep}
Farshad Saberi-Movahed, Mohammad~K Ebrahimpour, Farid Saberi-Movahed, Monireh Moshavash, Dorsa Rahmatian, Mahvash Mohazzebi, Mahdi Shariatzadeh, and Mahdi Eftekhari.
\newblock Deep metric learning with soft orthogonal proxies.
\newblock {\em arXiv preprint arXiv:2306.13055}, 2023.

\bibitem{roth2022non}
Karsten Roth, Oriol Vinyals, and Zeynep Akata.
\newblock Non-isotropy regularization for proxy-based deep metric learning.
\newblock In {\em Proceedings of the IEEE/CVF Conference on Computer Vision and Pattern Recognition}, pages 7420--7430, 2022.

\bibitem{mohan2020moving}
Deen~Dayal Mohan, Nishant Sankaran, Dennis Fedorishin, Srirangaraj Setlur, and Venu Govindaraju.
\newblock Moving in the right direction: A regularization for deep metric learning.
\newblock In {\em Proceedings of the IEEE/CVF Conference on Computer Vision and Pattern Recognition}, pages 14591--14599, 2020.

\bibitem{zhang2023threshold}
Qin ZHANG, Linghan Xu, Jun Fang, Qingming Tang, Ying~Nian Wu, Joseph Tighe, and Yifan Xing.
\newblock Threshold-consistent margin loss for open-world deep metric learning.
\newblock In {\em The Twelfth International Conference on Learning Representations}, 2024.

\bibitem{kim2023hier}
Sungyeon Kim, Boseung Jeong, and Suha Kwak.
\newblock Hier: Metric learning beyond class labels via hierarchical regularization.
\newblock In {\em Proceedings of the IEEE/CVF Conference on Computer Vision and Pattern Recognition}, pages 19903--19912, 2023.

\bibitem{kirchhof2022non}
Michael Kirchhof, Karsten Roth, Zeynep Akata, and Enkelejda Kasneci.
\newblock A non-isotropic probabilistic take on proxy-based deep metric learning.
\newblock In {\em European Conference on Computer Vision}, pages 435--454. Springer, 2022.

\bibitem{zhao2021towards}
Wenliang Zhao, Yongming Rao, Ziyi Wang, Jiwen Lu, and Jie Zhou.
\newblock Towards interpretable deep metric learning with structural matching.
\newblock In {\em Proceedings of the IEEE/CVF International Conference on Computer Vision}, pages 9887--9896, 2021.

\bibitem{venkataramanan2021takes}
Shashanka Venkataramanan, Bill Psomas, Ewa Kijak, Laurent Amsaleg, Konstantinos Karantzalos, and Yannis Avrithis.
\newblock It takes two to tango: Mixup for deep metric learning.
\newblock {\em arXiv preprint arXiv:2106.04990}, 2021.

\bibitem{ko2021learning}
Byungsoo Ko, Geonmo Gu, and Han-Gyu Kim.
\newblock Learning with memory-based virtual classes for deep metric learning.
\newblock In {\em Proceedings of the IEEE/CVF International Conference on Computer Vision}, pages 11792--11801, 2021.

\bibitem{barbany2024procsim}
Oriol Barbany, Xiaofan Lin, Muhammet Bastan, and Arnab Dhua.
\newblock Procsim: Proxy-based confidence for robust similarity learning.
\newblock In {\em Proceedings of the IEEE/CVF Winter Conference on Applications of Computer Vision}, pages 1308--1317, 2024.

\bibitem{ren2024learning}
Li~Ren, Chen Chen, Liqiang Wang, and Kien Hua.
\newblock Learning semantic proxies from visual prompts for parameter-efficient fine-tuning in deep metric learning.
\newblock {\em arXiv preprint arXiv:2402.02340}, 2024.

\bibitem{zheng2021deep}
Wenzhao Zheng, Borui Zhang, Jiwen Lu, and Jie Zhou.
\newblock Deep relational metric learning.
\newblock In {\em Proceedings of the IEEE/CVF International Conference on Computer Vision}, pages 12065--12074, 2021.

\bibitem{wang2023deep}
Chengkun Wang, Wenzhao Zheng, Junlong Li, Jie Zhou, and Jiwen Lu.
\newblock Deep factorized metric learning.
\newblock In {\em Proceedings of the IEEE/CVF Conference on Computer Vision and Pattern Recognition}, pages 7672--7682, 2023.

\bibitem{zheng2021deep2}
Wenzhao Zheng, Chengkun Wang, Jiwen Lu, and Jie Zhou.
\newblock Deep compositional metric learning.
\newblock In {\em Proceedings of the IEEE/CVF Conference on Computer Vision and Pattern Recognition}, pages 9320--9329, 2021.

\bibitem{zabihzadeh2024ensemble}
Davood Zabihzadeh, Zahraa Alitbi, and Seyed~Jalaleddin Mousavirad.
\newblock Ensemble of loss functions to improve generalizability of deep metric learning methods.
\newblock {\em Multimedia Tools and Applications}, 83(7):21525--21549, 2024.

\bibitem{li2023robust}
Xinyue Li, Jian Wang, Wei Song, Yanling Du, and Zhixiang Liu.
\newblock Robust calibrate proxy loss for deep metric learning.
\newblock {\em arXiv preprint arXiv:2304.09162}, 2023.

\bibitem{ren2024towards}
Li~Ren, Chen Chen, Liqiang Wang, and Kien Hua.
\newblock Towards improved proxy-based deep metric learning via data-augmented domain adaptation.
\newblock {\em arXiv preprint arXiv:2401.00617}, 2024.

\bibitem{furusawamean}
Takuya Furusawa.
\newblock Mean field theory in deep metric learning.
\newblock 2024.

\bibitem{imgaug}
Alexander~B. Jung, Kentaro Wada, Jon Crall, Satoshi Tanaka, Jake Graving, Christoph Reinders, Sarthak Yadav, Joy Banerjee, Gábor Vecsei, Adam Kraft, Zheng Rui, Jirka Borovec, Christian Vallentin, Semen Zhydenko, Kilian Pfeiffer, Ben Cook, Ismael Fernández, François-Michel De~Rainville, Chi-Hung Weng, Abner Ayala-Acevedo, Raphael Meudec, Matias Laporte, et~al.
\newblock {imgaug}.
\newblock \url{https://github.com/aleju/imgaug}, 2020.
\newblock Online; accessed 01-Feb-2020.

\bibitem{liu2017sphereface}
Weiyang Liu, Yandong Wen, Zhiding Yu, Ming Li, Bhiksha Raj, and Le~Song.
\newblock Sphereface: Deep hypersphere embedding for face recognition.
\newblock In {\em Proceedings of the IEEE conference on computer vision and pattern recognition}, pages 212--220, 2017.

\bibitem{huang2008labeled}
Gary~B Huang, Marwan Mattar, Tamara Berg, and Eric Learned-Miller.
\newblock Labeled faces in the wild: A database forstudying face recognition in unconstrained environments.
\newblock In {\em Workshop on faces in'Real-Life'Images: detection, alignment, and recognition}, 2008.

\bibitem{sengupta2016frontal}
Soumyadip Sengupta, Jun-Cheng Chen, Carlos Castillo, Vishal~M Patel, Rama Chellappa, and David~W Jacobs.
\newblock Frontal to profile face verification in the wild.
\newblock In {\em 2016 IEEE winter conference on applications of computer vision (WACV)}, pages 1--9. IEEE, 2016.

\bibitem{moschoglou2017agedb}
Stylianos Moschoglou, Athanasios Papaioannou, Christos Sagonas, Jiankang Deng, Irene Kotsia, and Stefanos Zafeiriou.
\newblock Agedb: the first manually collected, in-the-wild age database.
\newblock In {\em proceedings of the IEEE conference on computer vision and pattern recognition workshops}, pages 51--59, 2017.

\bibitem{zheng2017cross}
Tianyue Zheng, Weihong Deng, and Jiani Hu.
\newblock Cross-age lfw: A database for studying cross-age face recognition in unconstrained environments.
\newblock {\em arXiv preprint arXiv:1708.08197}, 2017.

\bibitem{zheng2018cross}
Tianyue Zheng and Weihong Deng.
\newblock Cross-pose lfw: A database for studying cross-pose face recognition in unconstrained environments.
\newblock {\em Beijing University of Posts and Telecommunications, Tech. Rep}, 5:7, 2018.

\bibitem{demoor2024realignedsoftmaxwarpingdeep}
Michael~G. DeMoor and John~J. Prevost.
\newblock Preprint: Realigned softmax warping for deep metric learning, 2024.

\end{thebibliography}
}


\newpage

\appendices
\section{Section 3 TLDR}
\label{sec:A}
Consider the rearranged Cross-Entropy loss with a single sample \(e_i\) and 2 classes (with a Euclidean distance inserted):
\begin{equation}
    CE = \log \Big(1 + \exp(||e_i - p_{c}|| - ||e_i - p_{c^{\prime}}||)\Big) 
\end{equation}
where \(p_c\) is the ground-truth proxy and \(p_{c^\prime}\) is the opposite class proxy. The optimization landscape for this loss can be warped to boost separability and preserve compactness by applying functions to \(||e_i - p_c||\) and \(||e_i - p_{c^\prime}||\).
\begin{equation}
    CE = \log \Big(1 + \exp\big(f_1(||e_i - p_{c}||) - f_2(||e_i - p_{c^{\prime}}||)\big)\Big)
\end{equation}
First, the extrema for this loss can be constrained entirely to \(L_p\) (the line intersecting \(p_c\) and \(p_{c^\prime}\)) by restricting \(f_1\) and \(f_2\) to be non-decreasing. From there, a separability-enhancing (and compactness-preserving) point of attraction can be formed at a point \(\alpha\) down this line by flipping the derivatives of \(f_1\) and \(f_2\) at \(\alpha\). See the Loss Interpretation in Appendix \ref{sec:B} for further insight. This boosts separability by creating a new point of attraction \emph{away} from both proxies 
(under normal circumstances proxies would serve as approximate ``cluster centers" for each class of embeddings to be attracted to).
And it preserves compactness by keeping the embeddings from moving ``too far" away and spreading.  

\begin{figure*}[t]
    \centering
    \subfloat[\(\Longrightarrow\)\label{fig:Lemma1-Illustrations-Necessary}]{
        \includegraphics[width=0.425\textwidth, height=5.50cm]{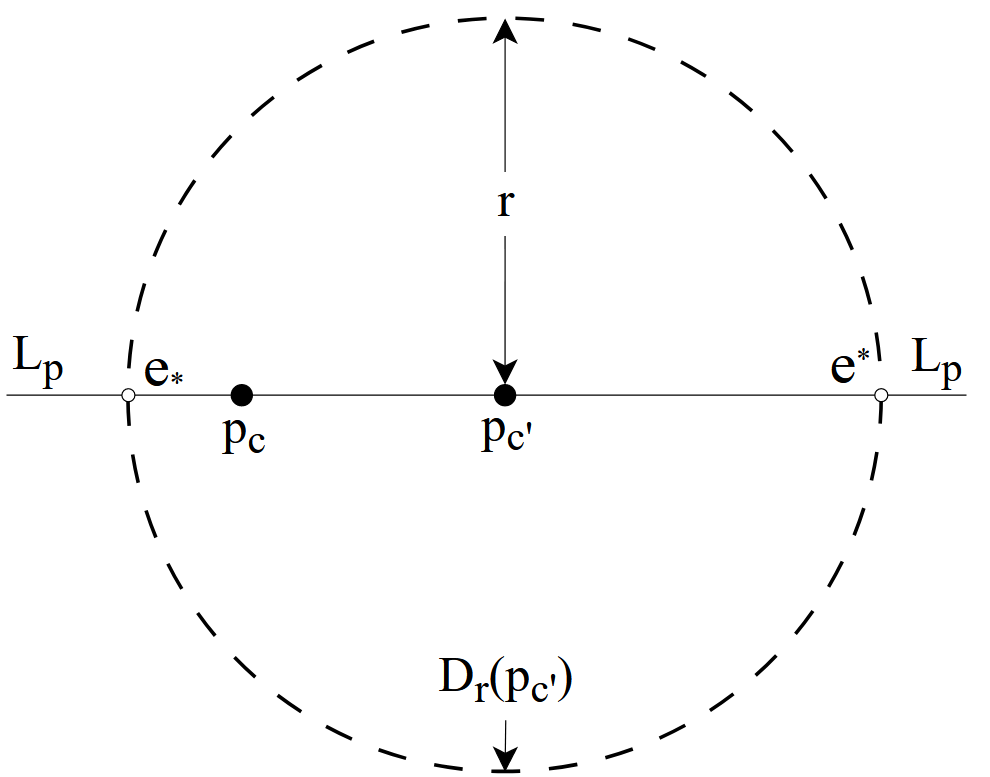}
    }
    \hspace{0.02\textwidth}
    \subfloat[\(\Longleftarrow\)\label{fig:Lemma1-Illustrations-Sufficient}]{
        \includegraphics[width=0.425\textwidth, height=5.50cm]{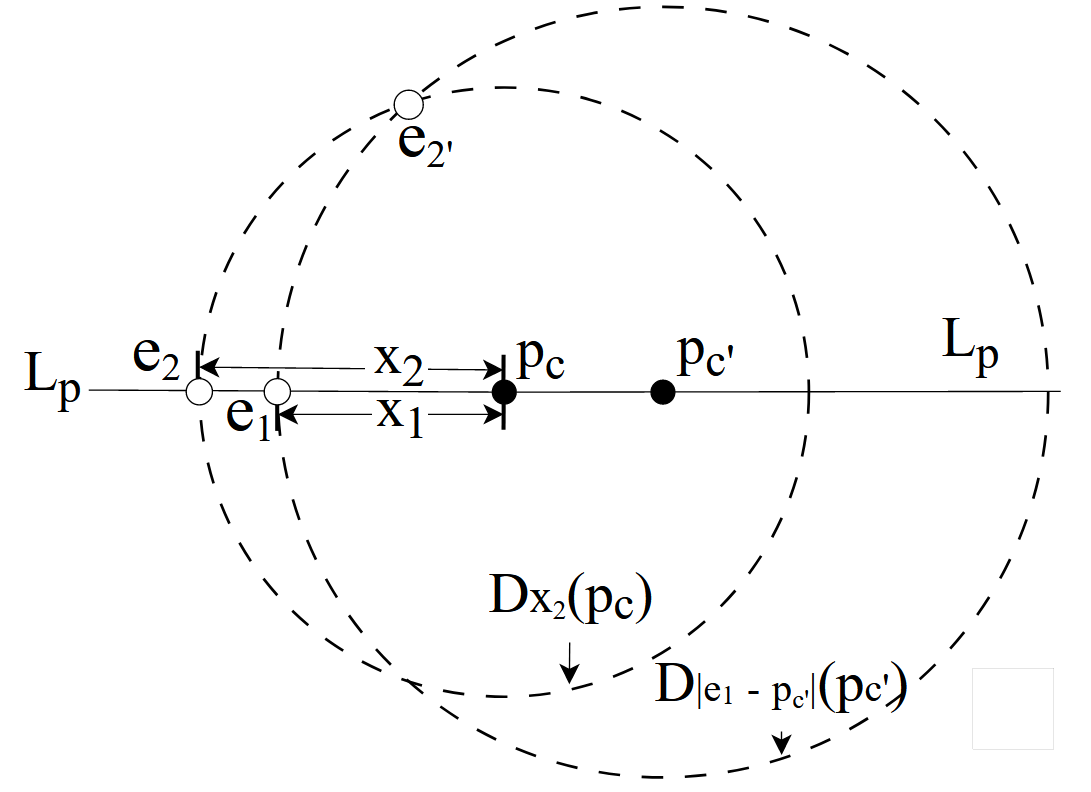}
    }
    \caption{Lemma 3.1 Illustration}    
    \label{fig:Lemma1-Illustrations}
\end{figure*}

\section{Further Explanation}
\label{sec:B}
In this appendix, we aim to provide some intuitive, less formal, and simplified explanations of the arguments presented in Lemma \ref{lemma:Lp-Extrema} and Proposition \ref{prop:Minimum-Creation}. We also further discuss their implications. For further simplicity, in addition to the binary class setting, we constrain our discussion to 2 dimensions. Note that:
\begin{equation}
    \arg\min_z \log\big(1 + \exp(z)\big) = \arg\min_z z.
\end{equation}
So the minimums in Eq. \ref{eq:Function-Softmax} are equivalent to the minimums of just:
\begin{equation}
    f(||e - p_c||) - f(||e - p_{c^\prime}||)
    \label{eq:Exponential-Softmax-Term}
\end{equation} 
This does not mean that Eq. \ref{eq:Exponential-Softmax-Term} has an identical optimization landscape to Eq. \ref{eq:Function-Softmax}. It just means they both share minimums at the exact same locations. The same goes for maximums. 

\subsection{Overview of Lemma 3.1}
\label{subsec:Lemma3.1-Overview}
\noindent We have the ground-truth proxy \(p_c\) and the opposite-class proxy \(p_{c^\prime}\). 
Disregarding \(f\), consider the optimization of:
\begin{equation}
    ||e - p_c|| - ||e - p_{c\prime}||
    \label{eq:d1-d2-Exponential-Term}
\end{equation}

To simplify it further, consider only the points in \(\mathbb{R}^2\) where \(r = ||e - p_{c^\prime}||\) is constant. These points form a ``disk" around \(p_{c^\prime}\) as shown in Fig. \ref{fig:Lemma1-Illustrations-Necessary}. Within this disk, the minimum of Eq. \ref{eq:d1-d2-Exponential-Term} is determined entirely by wherever \(||e - p_c||\) is shortest. 
Referring to Fig. \ref{fig:Lemma1-Illustrations-Necessary}, this point is \(e_*\). And it will always be \(e_*\) regardless of what circle you draw around \(p_{c^\prime}\).
The same can be said for the maximum. \(||e - p_c||\) will always be largest at \(e^*\). The same relationship also holds for all points traversing the circle ``between" \(e_*\) and \(e^*\) (going from smallest to largest). If \(f\) is monotone non-decreasing, then this fact is still true for \(f\). Additionally, this is also true irregardless of the choice of proxies \(p_c\) and \(p_{c^\prime}\).

\vspace{0.25cm}
\noindent\textit{\(\Longrightarrow\): If \(f\) is monotone then \(e_*\) (\(e^*\)) is the only minimum (maximum) of Eq. \ref{eq:Exponential-Softmax-Term} within disk r for any \(r\) and for any \(p_c\).} 
\vspace{0.25cm}

\noindent This means that, \emph{\underline{regardless}} of the proxies, any local or global extrema of Eq. \ref{eq:Function-Softmax}/\ref{eq:Exponential-Softmax-Term} will fall on \(L_p\) as long as \(f\) is monotone. If there's an extrema \(e_0\) not on \(L_p\), draw a \(||e_0 - p_{c^\prime}||\) circle through it to make a contradiction. 

Conversely, it can be shown that only monotone functions are capable of doing this (forcing any/all extrema onto \(L_p\)). If \(f\) is not monotone, then ``somewhere" it is decreasing. Let \(x_1 < x_2 \in \mathbb{R}\) denote points where \(f(x_1) > f(x_2)\). Referring to Fig. \ref{fig:Lemma1-Illustrations-Sufficient}, 
if we choose \(e_1, e_2\) so that \(x_1 = ||e_1 - p_c||\) and \(x_2 = ||e_2 - p_c||\) then \(f(||e_1 - p_c||) > f(||e_2 - p_c||)\). Let \({e_2}^\prime\) be one of the points in the intersection of the \(p_{c^\prime}\) and \(p_c\) circles. Note that since \({e_2}^\prime\) and \(e_2\) are both points within the \(p_c\) circle, \(f(||{e_2}^\prime - p_c||) = f(||e_2 - p_c||) < f(||e_1 - p_c||)\). A contradiction can then be formed by comparing \(e_1\) with \({e_2}^\prime\). Both \(e_1\) and \({e_2}^\prime\) are points within the \(p_{c^\prime}\) circle, but \(e_1\) (\(e_*\) here) is not the minimum of Eq. \ref{eq:Exponential-Softmax-Term} within it.

\vspace{0.25cm}
\noindent\textit{\(\Longleftarrow\): Only monotone functions are capable of doing this.}

\begin{figure}[t]
    \centering
    \includegraphics[width=\columnwidth, height=2.5cm]{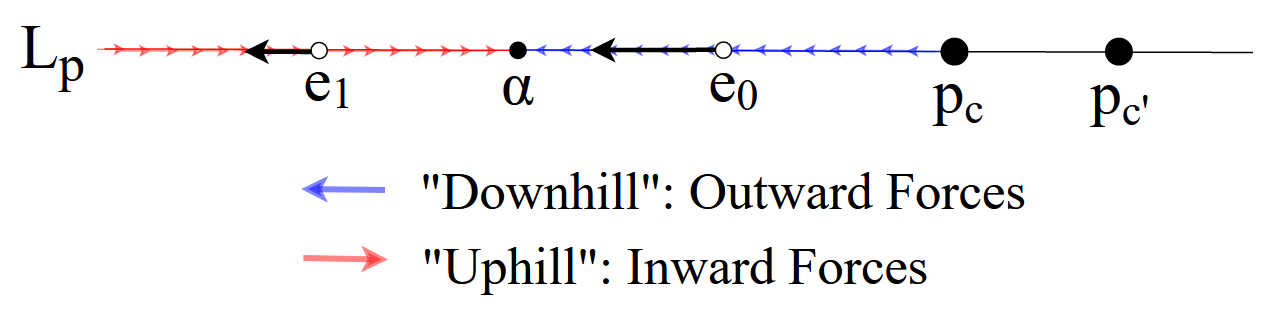}
    \caption{Proposition 3.2 Illustration}   
    \label{fig:Prop1-2-Illustrations}
\end{figure}

\subsection{Overview of Proposition 3.2}
\label{subsec:Prop3.2-Overview}
\noindent By the Lemma we only need to focus on the 1-dimensional line \(L_p\). Referring to Fig. \ref{fig:Prop1-2-Illustrations}, Prop. \ref{prop:Minimum-Creation} considers all the points on \(L_p\) to the ``left" of \(p_c\) (in blue/red). 
Sliding a point \(e\) down this portion of the line, if \(\frac{df_1}{dt} < \frac{df_2}{dt}\), then \(f_2\) is increasing \emph{faster} than \(f_1\). 
Therefore, \(f_1 - f_2\) is decreasing and \(e\) is going \emph{downhill} (e.g. \(e_0\) in Fig. \ref{fig:Prop1-2-Illustrations}). 
Similarly, if \(\frac{df_1}{dt} > \frac{df_2}{dt}\), then \(f_1 - f_2\) is increasing and \(e\) is going \emph{uphill} (e.g. \(e_1\) in Fig. \ref{fig:Prop1-2-Illustrations}). By reversing these inequalities at some point \(\alpha\), \(e\) is \emph{switched} from going downhill (decreasing) to going uphill (increasing). That is, a minimum is created at the point of reversal \(\alpha\).

\vspace{0.25cm}
\noindent\textit{Reversing the \(\frac{df_1}{dt} < \frac{df_2}{dt}\) inequality at a point \(\alpha\) creates a minimum at \(\alpha\) in Eq. \ref{eq:Function-Softmax-Split}.}

\subsection{Interpretations of Proposed Loss (Re-Stated)}
\label{subsec:Loss-Interpretations-ReStated}
\noindent For our loss proposed in Eq. \ref{eq:Warp-Function}/\ref{eq:Multi-Class-Warp}, \(k_1, k_2\), and \(\alpha\) are all derived with respect to the following insight which builds off of Prop. 3.2. 
For an embedding vector \(e\), the strength of the attractive force towards any minimum is directly related to the difference between \(\frac{df_1}{dt}\) and \(\frac{df_2}{dt}\). Roughly speaking, if \(\frac{df_1}{dt} < \frac{df_2}{dt}\), \(e\) will travel downhill to move outwards. If \(\frac{df_1}{dt} > \frac{df_2}{dt}\), \(e\) will need to travel uphill to move outwards. If \(\frac{df_1}{dt} << \frac{df_2}{dt}\), \(e\) will strongly accelerate outwards. If \(\frac{df_1}{dt} >> \frac{df_2}{dt}\), \(e\) will be strongly anchored inward.

Therefore, \(\frac{df_1}{dt} < \frac{df_2}{dt}\) boosts separability. 
And \(\frac{df_1}{dt} > \frac{df_2}{dt}\) boosts compactness. Since \(f_2(t) = t, \frac{df_1}{dt} = k < 1\) boosts separability and \(k > 1\) compactness.   
As such, \(k_1 < 1\) determines the ``strength" of the outward forces. \(k_2 > 1\) determines the ``strength" of the inward forces. 
\(\alpha\) is the effective point of attraction (the minimum). It determines ``where" to apply the inward and outward forces respectively.
\(\alpha = 0 \; (\infty)\) implies inward (outward) forces everywhere. Interestingly, our loss (without \(\Delta\)) actually has the opposite effect of traditional margin-based losses on the \([0,\alpha)\) interval. This is because \(k_1\) artificially handicaps the loss by decreasing the distance between an embedding and it's ground-truth proxy. However, this is necessary in order to boost separability. \(\Delta\)'s first purpose in Eq. \ref{eq:Warp-Function} is to simply correct this handicap. But it can be increased even more to help boost the loss further (thus serving more like a margin in the traditional sense).

\newpage
\section{Proofs of Propositions}
\label{sec:C}

\subsection{Lemma 3.1}
\label{subsec:Lemma3.1-Proof}
\begin{flushleft}
\textit{Given both \(p_c\) and \(p_{c^\prime}\), \(f\) is monotone if and only if for any \(r \in \mathbb{R}\),  \(e_*\) and \(e^*\) are respectively the only minimum and maximum of Eq. \ref{eq:Function-Softmax} within the r-disk \(D_r(p_{c^{\prime}})\).}
\end{flushleft}

\vspace{0.25cm}
\noindent Proof \(\Rightarrow\): The goal is to minimize \(f(||e - p_c||) - f(||e -p_{c^{\prime}}||)\).
In 2 dimensions, consider the r-disk \(D_r(p_{c^{\prime}})\).
When \(f = ||\cdot||\), the minimum is determined entirely by \(||e - p_c||\) (Fig. \ref{fig:Lemma1-Illustrations-Necessary}). 
Given \(e \in D_r(p_{c^{\prime}})\), let \(r = ||e - p_{c^\prime}||\), \(d = ||p_c - p_{c^\prime}||\), and \(\theta\) denote the angle between them. By the Law of Cosines:
\begin{equation}
    ||e - p_c||^2 = r^2 + d^2 + 2rd\cos(\theta)
\end{equation}
Because \(r\) and \(d\) are constant \(\forall e \in D_r(p_{c^{\prime}})\), \(||e - p_c||\) is a function of \(\theta\) alone. And since \(\cos(\theta)\) is monotonically decreasing from \(0^\circ\) (\(e^*\)) to \(180^\circ\) (\(e_*\)), so is \(||e - p_c||\). As such, the only minimum (maximum) of \(||e - p_c||\) within \(D_r(p_{c^{\prime}})\) is \(e_*\) (\(e^*\)). By extension, the same follows for the monotone function \(f\). Within the n-dimensional setting, this argument can be applied on the 2d plane intersecting \(p_c\), \(p_{c^\prime}\), and \(e\) to show that it still holds true regardless of dimensionality. 

\vspace{0.25cm}
\noindent Proof \(\Leftarrow\):
Let \(L_p\) serve as the x-axis and \(p_c < p_{c^\prime}\) (with no loss to generality). 
If \(f\) is not monotone, then \(\exists e_1, e_2 \in L_p\) such that \(e_2 < e_1 < p_c\) and \(|e_1 - p_c| < |e_2 - p_c|\) but \(f(|e_1 - p_c|) > f(|e_2 - p_c|)\). 
Letting \(x_1 = |e_1 - p_{c^\prime}|\) and \(x_2 = |e_2 - p_c|\), respectively,
note that \(D_{x_1}(p_{c^\prime})\) and \(D_{x_2}(p_c)\) overlap (Fig. \ref{fig:Lemma1-Illustrations-Sufficient}). Let \({e_2}^\prime \in \{D_{x_1}(p_{c^\prime}) \cap D_{x_2}(p_c)\}\). Now since \(||e_2 - p_c|| = ||{e_2}^\prime - p_c||\),
\begin{equation}
f(||e_1 - p_c||) > f(||e_2 - p_c||) = f(||{e_2}^\prime - p_c||)
\end{equation}
This is a contradiction because both \(e_1\) and \({e_2}^\prime\) are points within disk \(x_1\) but \(e_1\) is not the minimum.


\subsection{Proposition 3.2}
\label{subsec:Prop3.2-Proof}
\begin{flushleft}
\textit{Assuming \(f_{1,2} \in C^2\) \(\mathfrak{(a.e.)}\) and monotone, then \(\forall e \in B_{||p_c - p_{c^{\prime}}||}(p_{c^{\prime}}) \setminus L_p\) such that \(||e - p_c|| < ||e - p_{c^\prime}||\), reversing \(\frac{df_1}{dt} < \frac{df_2}{dt}\) at \(e\) is equivalent to creating a minimum in Eq. \ref{eq:Function-Softmax-Split} at \(e\).
}
\end{flushleft}
Note: \textit{\(\mathfrak{(a.e.)}\)} stands for ``almost everywhere" and and is used here to denote basically all points except potentially \(e\) itself.

Notation reminder:
for non-negative functions \(f_1(t), f_2(t)\), \(\frac{df_1}{dt}, \frac{df_2}{dt}\) is an abbreviation for \(\frac{df_1(t)}{dt}|_{t=||e - p_c||}\) and \(\frac{df_2(t)}{dt}|_{t=||e - p_{c^{\prime}}||}\), respectively.
Additionally, \(f_1(e)\) and \(f_2(e)\) refer to \(f_1(||e - p_c||)\) and \(f_2(||e - p_{c^{\prime}}||)\), respectively. Again, by the lemma, any minimum is \(\in L_p\), so we restrict our consideration to only these points. Again, without any loss to generality, let \(L_p\) be the x-axis and \(p_c < p_{c^{\prime}} = 0\).

\vspace{0.25cm}
\noindent Proof \(\Rightarrow\): Let \(I = \{e \in B_{||p_c - p_{c^{\prime}}||}(p_{c^{\prime}}) \setminus L_p \,; \, ||e - p_c|| < ||e - p_{c^\prime}||\}\) \(= \{e \in L_p \,; \, e < p_c\}\). 
Denote the point of reversal as \(e_1 \in I\) and select \(e_0, e_2 \in I\) so that \(e_0 < e_1 < e_2 < p_c\). 
On the \([e_0, e_1)\) interval \(\frac{df_1}{dt} > \frac{df_2}{dt}\). Therefore, \(e_0\) is not the minimum since \(f_1(e_0) - f_1(e_1) > f_2(e_0) - f_2(e_1) \iff f_1(e_0) - f_2(e_0) > f_1(e_1) - f_2(e_1)\). Similarly \(e_2\) is not the minimum. Therefore \(e_1\) is the minimum since \(e_0\) and \(e_2\) are arbitrary. 

\vspace{0.25cm}
\noindent Proof \(\Leftarrow\): This is trivial.

\section{Reality Check}
\label{sec:D}

In this appendix, we run experiments inspired by the work of \cite{musgrave2020metric}. The authors make a number of criticisms of modern DML papers and introduce three new metrics to measure the performance of DML losses.

\subsection{\textbf{Implementation and Setup}}
We select three of the top performing losses in \cite{musgrave2020metric} for each dataset, the two proxy-based losses most closely related to our own (ProxyNCA++ and Proxy-Anchor), and for CUB and CARS one of the more recently published loss functions HIST Loss \cite{lim2022hypergraph} and rerun all experiments for these losses alongside our own. 
We employ the same base model and optimizer universally across all loss functions and add max+average pooling and Layer-Normalization to all of them. For CUB and CARS we use a ResNet50 \cite{he2016deep} backbone and for SOP we use BNInception \cite{ioffe2015batch}. RMSProp is adopted as the optimizer.  
We also use the same crop size, augmentation setup, and embedding dimensionality across all loss functions.

A full reality check would involve evaluating all recent losses in the field under the same setting. This is no small undertaking which is further made difficult by the large diversity in approaches. 
As such, since this is not a reality check paper and this is not our focus, we leave such an endeavor worthy of its own paper to the future. Instead, we opt to select one of the more ``recent" losses and proceed accordingly. 
We select HIST Loss for two primary reasons: Ease of code integration, and it performs the best among the losses in Table 1 (main paper) that don't utilize additional model/data components in their setup.

\subsubsection{\textbf{Training:}}
For each dataset, we run 50 iterations of Bayesian hyperparameter optimization for each loss function. As in \cite{musgrave2020metric}, each dataset is split 50/50 trainval/test and then for every iteration we apply 4-fold Cross-Validation on the trainval split.
Every model is trained in 2-phases. In the normal phase the model is trained until validation accuracy plateaus. In the followup phase we lower the learning rate(s), adjust the parameters to followup values, and continue until validation accuracy plateaus once more. The followup parameters are their own set of hyyperparameters.

\subsubsection{\textbf{Evaluation:}}
We follow the same procedure as in \cite{musgrave2020metric} for evaluation under the separated 128-dim (aggregate) and concatenated 512-dim (ensemble) categories. We use the three introduced metrics: MAP@R, RP, and P@1. Unlike before, hyperparameters are optimized with respect to MAP@R instead of P@1 (equivalent to R@1 in Table 1 Section 4 in the main paper).
We do 10 runs with the final hyperparameters and report the 95\% confidence interval for each metric and category. The results are displayed in Tab. \ref{tab:CUB-Reality-Check}, \ref{tab:CARS-Reality-Check}, and \ref{tab:SOP-Reality-Check}.

\subsection{\textbf{Another Category}}
Due to cross validation, no individual model trained makes use of the full trainval split. As such, none of the trained models make use of all the information available within the dataset. Furthermore, no models are trained end-to-end on a standard 512-dim space. To investigate any potential impacts resulting from these setting choices, we 
choose to evaluate under an additional "Complete" procedure with the goal of evaluating loss functions more holistically. 

\subsubsection{\textbf{Procedure:}}
To evaluate this category, we use the same optimal hyperparameters used for the other categories and change the embedding dimensionality to 512. Since there is no longer a validation set to determine when to stop training, we estimate the optimal number of epochs for both training phases by averaging the number of total and followup epochs off the eval runs used for the other categories. We then round up to the nearest 5th epoch and train the model end-to-end using these values. The rest of the setup is the same. The results are also displayed in Tab. \ref{tab:CUB-Reality-Check}, \ref{tab:CARS-Reality-Check}, and \ref{tab:SOP-Reality-Check}. 

\subsection{\textbf{Comparison}}
Tab. \ref{tab:CUB-Reality-Check} displays the results on CUB. Our loss outperforms all others on the MAP@R and RP metrics for both the Aggregate and EnsembleCategories. For the Complete Category it performs the best on 2 of the 3 metrics, and comes in second on the other one. 
Tab. \ref{tab:CARS-Reality-Check} displays the results on Cars. Our loss achieves its worst comparative results on this dataset, being outperformed by Multi-Similarity and CosFace under the Ensemble and Multi-Similarity and ProxyNCA++ under the Aggregate. However, it is still competitive, and actually performs the best on the P@1 metric on the Aggregate category. Furthermore, it performs the best on all metrics under the Complete Category. 
Results for SOP can be found in Tab. \ref{tab:SOP-Reality-Check}.
While outperformed under the Aggregate, our loss (along with ProxyNCA++) actually surpasses ProxyNCA when training under the full dataset on a complete 512-dim model. Our loss performs the best under the Complete results for the MAP@R and RP metrics, and second best on every other metric/category. Overall, we achieve competitive/superior results on these datasets and overall strongly outperform our baseline (closest related method) ProxyNCA++.

\subsection{\textbf{Further Discussion}}
The Complete results offer a more holistic perspective by portraying the state of a single model trained end-to-end with the entire training set (without any cross validation). The results also illustrate the difference between artificially constructing a 512-dim space via an ensemble vs training a single model on one end-to-end. Various losses are impacted differently. Most of them scale well with the increased dimensions and full training set. Some overtake others with the changes (e.g. CosFace on CUB). Unsurprisingly, the Complete results are better than the Aggregate (with the exception of the interesting case with MultiSimilarity).
\begin{table*}[th]
    \caption{Results on CUB-200-2011}
    \label{tab:CUB-Reality-Check}
    \vspace{0.1in}
\begin{adjustbox}{width=1.0\textwidth,center=\textwidth}
\renewcommand{\arraystretch}{1.75}
\huge
\begin{tabular}{ l | c  c  c | c  c  c | c  c  c } 
     \hline
     & \multicolumn{3}{|c|}{Complete (512-dim)} & \multicolumn{3}{|c|}{Ensemble (512-dim)} & \multicolumn{3}{|c}{Aggregate (128-dim)}\\
     \hline
     Method & MAP@R & RP & P@1 & MAP@R & RP & P@1 & MAP@R & RP & P@1\\
     \hline
     Contrastive \cite{hadsell2006dimensionality} & 23.63 \(\pm\) 0.24 & 34.77 \(\pm\) 0.23 & 63.0 \(\pm\) 0.63 & 25.90 \(\pm\) 0.41 & 36.80 \(\pm\) 0.40 & 66.25 \(\pm\) 0.41 & 21.27 \(\pm\) 0.33 & 32.24 \(\pm\) 0.33 & 59.53 \(\pm\) 0.41\\
     
     CosFace \cite{wang2018cosface} & \textbf{26.89 \(\pm\) 0.22} & \underline{37.69 \(\pm\) 0.21} & \underline{66.80 \(\pm\) 0.16} & \underline{26.95 \(\pm\) 0.15} & \underline{37.93 \(\pm\) 0.16} & 67.07 \(\pm\) 0.18 & 22.72 \(\pm\) 0.12 & 33.77 \(\pm\) 0.12 & \underline{61.57 \(\pm\) 0.25}\\

     ArcFace \cite{deng2019arcface} & 24.59 \(\pm\) 0.36 & 35.30 \(\pm\) 0.38 & 63.64 \(\pm\) 0.38 & 26.63 \(\pm\) 0.14 & 37.60 \(\pm\) 0.13 & 66.77 \(\pm\) 0.33 & \underline{22.77 \(\pm\) 0.11} & 33.85 \(\pm\) 0.12 & 61.48 \(\pm\) 0.25\\

     PrAnchor \cite{kim2020proxy} & 25.74 \(\pm\) 0.21 & 36.55 \(\pm\) 0.23 & 64.94 \(\pm\) 0.32 & 25.72 \(\pm\) 0.28 & 36.71 \(\pm\) 0.28 & 65.90 \(\pm\) 0.38 & 22.23 \(\pm\) 0.19 & 33.28 \(\pm\) 0.19 & 60.90 \(\pm\) 0.31\\
     
     PrNCA++ \cite{teh2020proxynca++} & 25.96 \(\pm\) 0.41 & 36.91 \(\pm\) 0.42 & 65.56 \(\pm\) 0.36 & 26.82 \(\pm\) 0.23 & 37.82 \(\pm\) 0.22 & \underline{67.20 \(\pm\) 0.25} & 22.43 \(\pm\) 0.13 & 33.49 \(\pm\) 0.13 & 61.06 \(\pm\) 0.26\\

     HIST \cite{lim2022hypergraph} & 26.26 \(\pm\) 0.16 & 37.24 \(\pm\) 0.17 & 66.69 \(\pm\) 0.32 & 26.42 \(\pm\) 0.29 & 37.44 \(\pm\) 0.30 & 66.54 \(\pm\) 0.40 & 22.76 \(\pm\) 0.22 & \underline{33.87 \(\pm\) 0.23} & 61.54 \(\pm\) 0.33\\
     
     \rowcolor{lightgray} Warped-Softmax & \underline{26.76\(\pm\) 0.25} & \textbf{37.74 \(\pm\) 0.24} & \textbf{67.11 \(\pm\) 0.34} &  \textbf{28.47 \(\pm\) 0.29} & \textbf{39.35 \(\pm\) 0.26} & \textbf{69.63 \(\pm\) 0.38} & \textbf{23.01 \(\pm\) 0.21} & \textbf{34.09 \(\pm\) 0.21} & \textbf{62.4 \(\pm\) 0.26}\\
     \hline
\end{tabular}
\renewcommand{\arraystretch}{1}
\end{adjustbox}
\end{table*}

\begin{table*}[th]
    \caption{Results on CARS196}
    \label{tab:CARS-Reality-Check}
    \vspace{0.1in}
\begin{adjustbox}{width=1.0\textwidth,center=\textwidth}
\renewcommand{\arraystretch}{1.75}
\huge
\begin{tabular}{ l | c  c  c | c  c  c | c  c  c } 
     \hline
     & \multicolumn{3}{|c|}{Complete (512-dim)} & \multicolumn{3}{|c|}{Ensemble (512-dim)} & \multicolumn{3}{|c}{Aggregate (128-dim)}\\
     \hline
     Method & MAP@R & RP & P@1 & MAP@R & RP & P@1 & MAP@R & RP & P@1\\
     \hline
     Multi-Sim. \cite{wang2019multi} & 15.01 \(\pm\) 0.35 & 24.53  \(\pm\) 0.40 & 61.39 \(\pm\) 0.55 & \textbf{35.25  \(\pm\) 0.23} & \textbf{44.05  \(\pm\) 0.22} & \underline{89.55  \(\pm\) 0.17} & \textbf{25.22  \(\pm\) 0.22} & \textbf{35.30  \(\pm\) 0.23} & 79.52 \(\pm\) 0.44\\
     
     CosFace \cite{wang2018cosface} & \underline{29.36 \(\pm\) 0.82} & \underline{38.39  \(\pm\) 0.75} & \underline{86.67 \(\pm\) 0.75} & \underline{33.72  \(\pm\) 0.37} & \underline{42.55 \(\pm\) 0.27} & \textbf{90.13 \(\pm\) 0.28} & 22.64  \(\pm\) 0.28 & 32.5  \(\pm\) 0.22 & \underline{80.21  \(\pm\) 0.39}\\
 
     SoftTriple \cite{qian2019softtriple} & 28.25 \(\pm\) 0.41 & 38.12 \(\pm\) 0.37 & 84.79  \(\pm\) 0.41 & 32.16  \(\pm\) 0.23 & 41.50  \(\pm\) 0.20 & 88.52  \(\pm\) 0.23 &  22.00 \(\pm\) 0.24 & 32.26 \(\pm\) 0.22 & 78.03  \(\pm\) 0.44\\

     PrAnchor \cite{kim2020proxy} &  24.92 \(\pm\) 0.51 & 34.97  \(\pm\) 0.56 & 79.75  \(\pm\) 0.43 & 30.0  \(\pm\) 0.27 & 39.81  \(\pm\) 0.24 & 85.67 \(\pm\) 0.17 &  21.67 \(\pm\) 0.22 & 32.07 \(\pm\) 0.23 & 75.81  \(\pm\) 0.33\\
     
     PrNCA++ \cite{teh2020proxynca++} & 27.91  \(\pm\) 0.30 & 37.72  \(\pm\) 0.27 & 82.41 \(\pm\) 0.43 & 31.96  \(\pm\) 0.23 & 41.29  \(\pm\) 0.22 & 87.37 \(\pm\) 0.26 &  \underline{23.93 \(\pm\) 0.23} & \underline{34.08  \(\pm\) 0.23} & 78.78  \(\pm\) 0.27\\

     HIST \cite{lim2022hypergraph} & 28.19 \(\pm\) 0.26 & 37.66 \(\pm\) 0.27 & 84.5 \(\pm\) 0.27 & 31.17 \(\pm\) 0.28 & 40.56 \(\pm\) 0.25 & 88.25 \(\pm\) 0.20 & 22.28 \(\pm\) 0.23 & 32.56 \(\pm\) 0.22 & 78.48 \(\pm\) 0.27\\
     
     \rowcolor{lightgray} Warped-Softmax & \textbf{30.26 \(\pm\) 0.77} & \textbf{39.83 \(\pm\) 0.65} & \textbf{87.79 \(\pm\) 0.74} & 32.38 \(\pm\) 0.33 & 41.69  \(\pm\) 0.30 & 89.45 \(\pm\) 0.34 & 23.10  \(\pm\) 0.29 & 33.43 \(\pm\) 0.27 & \textbf{80.84 \(\pm\) 0.38}\\
     \hline
\end{tabular}
\renewcommand{\arraystretch}{1}
\end{adjustbox}
\end{table*}
\begin{table*}[th]
    \caption{Results on Stanford Online Products}
    \label{tab:SOP-Reality-Check}
    \vspace{0.1in}
\begin{adjustbox}{width=1.0\textwidth,center=\textwidth}
\renewcommand{\arraystretch}{1.75}
\huge
\begin{tabular}{ l | c  c  c | c  c  c | c  c  c } 
     \hline
     & \multicolumn{3}{|c|}{Complete (512-dim)} & \multicolumn{3}{|c|}{Ensemble (512-dim)} & \multicolumn{3}{|c}{Aggregate (128-dim)}\\
     \hline
     Method & MAP@R & RP & P@1 & MAP@R & RP & P@1 & MAP@R & RP & P@1\\
     \hline
     ProxyNCA \cite{movshovitz2017no} & 46.16  \(\pm\) 0.08 & 49.11  \(\pm\) 0.08 & 74.52  \(\pm\) 0.06 & \textbf{48.9  \(\pm\) 0.14} & \textbf{51.8  \(\pm\) 0.14} &\textbf{ 76.84  \(\pm\) 0.12} & \textbf{44.47  \(\pm\) 0.11} & \textbf{47.48  \(\pm\) 0.11} &\textbf{ 73.28  \(\pm\) 0.09}\\
     
     N. Softmax \cite{zhai2018classification} & 44.03  \(\pm\) 0.05 & 46.96  \(\pm\) 0.05 & 73.47  \(\pm\) 0.07 & 46.37  \(\pm\) 0.16 & 49.28  \(\pm\) 0.15 & 75.19  \(\pm\) 0.14 & 41.99  \(\pm\) 0.16 & 45.01  \(\pm\) 0.16 & 71.56  \(\pm\) 0.15\\

     Arcface \cite{deng2019arcface} & 42.96  \(\pm\) 0.06 & 45.83  \(\pm\) 0.06 & 73.00  \(\pm\) 0.09 & 47.78  \(\pm\) 0.16 & 50.66  \(\pm\) 0.17 & 76.39  \(\pm\) 0.13 & 41.93  \(\pm\) 0.02 & 44.88  \(\pm\) 0.17 & 71.74  \(\pm\) 0.15\\

     PrAnchor \cite{kim2020proxy} & 33.02  \(\pm\) 9.07 & 35.61  \(\pm\) 10.48 & 58.75  \(\pm\) 16.08 & 45.96  \(\pm\) 0.23 & 48.89  \(\pm\) 0.23 & 74.65  \(\pm\) 0.18 & 41.58  \(\pm\) 0.17 & 44.61  \(\pm\) 0.17 & 70.87  \(\pm\) 0.13\\
     
     PrNCA++ \cite{teh2020proxynca++} & \underline{46.26  \(\pm\) 0.08} & \underline{49.29  \(\pm\) 0.08} & \textbf{75.02  \(\pm\) 0.09} & 47.86  \(\pm\) 0.15 & 50.83  \(\pm\) 0.15 & 76.22  \(\pm\) 0.11 & 43.34  \(\pm\) 0.13 & 46.42  \(\pm\) 0.13 & 72.52  \(\pm\) 0.1\\
     
     \rowcolor{lightgray} Warped-Sotmax & \textbf{46.81  \(\pm\) 0.08} & \textbf{49.77  \(\pm\) 0.07} & \underline{74.89  \(\pm\) 0.07} &  \underline{48.45  \(\pm\) 0.18} & \underline{51.34  \(\pm\) 0.18} & \underline{76.41  \(\pm\) 0.13} & \underline{44.1  \(\pm\) 0.16} & \underline{47.19  \(\pm\) 0.15} & \underline{72.81  \(\pm\) 0.16}\\
     \hline
\end{tabular}
\renewcommand{\arraystretch}{1}
\end{adjustbox}
\end{table*}
\FloatBarrier
\noindent Generally speaking, the ensemble still performs better than training a single model on the full dataset, although by a smaller margin. Overall, our simple warped softmax example achieves competitive/superior performance on these datasets, and surpasses our baseline (ProxyNCA++) in 24 out of 27 categories.

\newpage
\begin{IEEEbiography}[{\includegraphics[width=0.98in,height=1.25in]{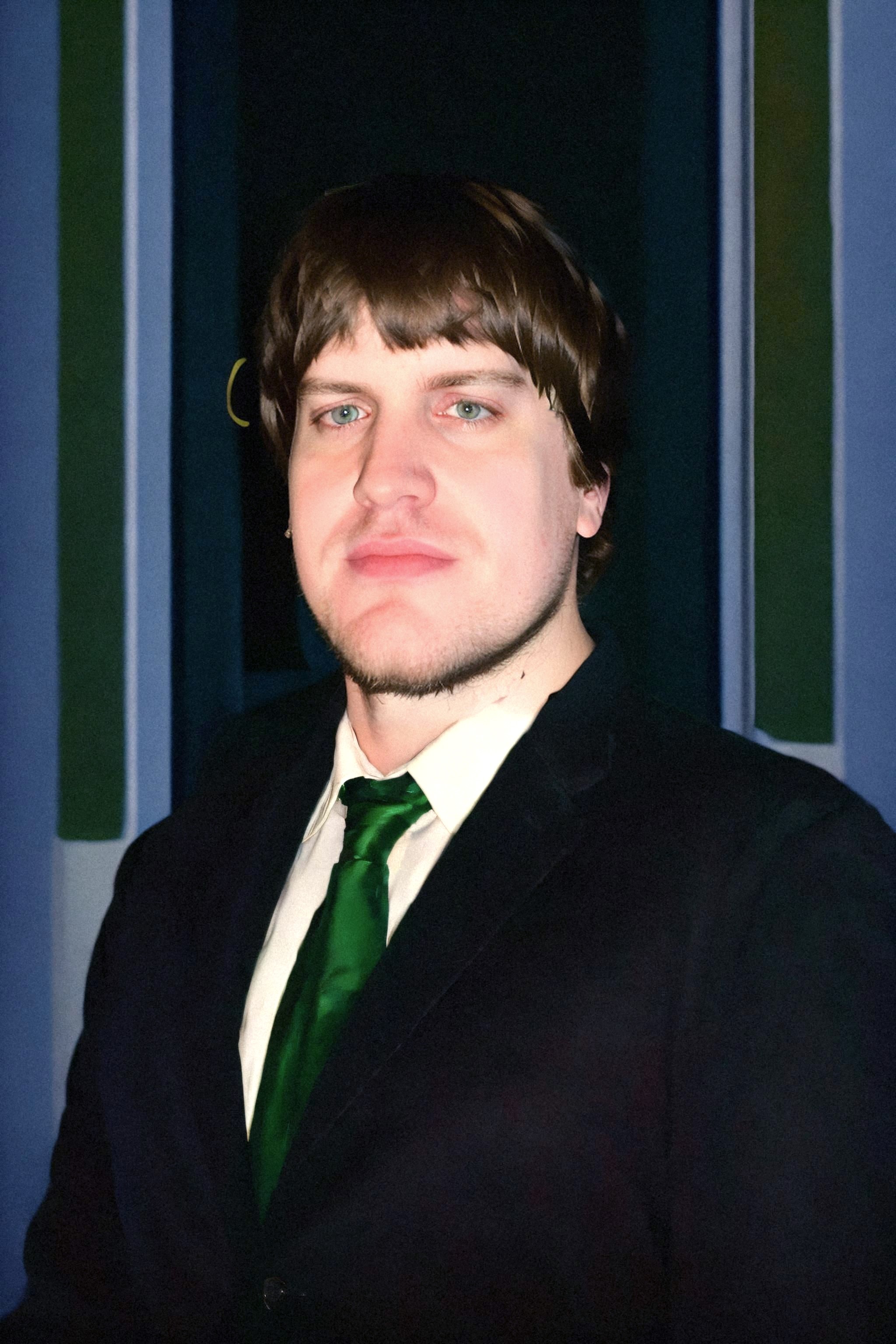}}]{Michael Goode DeMoor}
Michael G. DeMoor obtained a Bachelor of Science degree in both Electrical Engineering and Mathematics from the University of Texas at San Antonio (UTSA). He has since continued working in the Cloud Lab for Engineering Application Research (CLEAR) working on a number of various Deep Learning, IoT, and Computer Vision projects. He obtained a Master of Science in Computer Engineering in 2019 and is currently finishing up a doctoral degree. He is a member of the Open Cloud Institute (OCI). He is an active math enthusiast. And his current research interests include Metric Learning, Deep Metric Learning, Similarity Learning, Image Segmentation, and various other subdomains of Artificial Intelligence along with their applications across diverse fields of Engineering and Science.
\end{IEEEbiography}

\begin{IEEEbiography}
[{\includegraphics[width=1in,height=1.25in,clip,keepaspectratio]{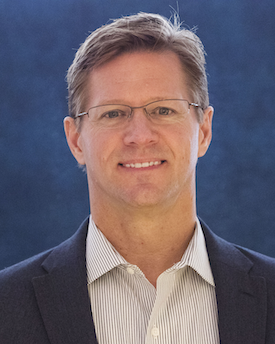}}]{John Jeffery Prevost}
Dr. John Jeffery Prevost attended Texas A\&M University and received an undergraduate degree. At the University of Texas at San Antonio he completed his masters and Ph.D. During his industry career, he worked in many different positions, serving in roles such as a chief consultant, Director of Information Systems, Director of Product Development, and Chief Technical Officer. He began his professional academic journey in 2013 as a professor of research. In 2015, he co-founded and became the Chief Research Officer and Assistant Director of the Open Cloud Institute, where he currently serves as its Executive Director. He currently serves as the Cloud Technology Endowed Associate Professor in the Electrical and Computer Engineering Department at UTSA. He currently also serves as the VP for Secure Cloud Architecture for the Cyber Manufacturing Innovation Institute. Working on behalf of the needs of his industry connections, he established the Graduate Certificate in Cloud Computing, which officially began in 2017. He remains an active consultant in areas of complex systems and cloud computing and continues to maintain strong ties with industry leaders. His is a member of Tau Beta Pi, Phi Kappa Phi and Eta Kappa Nu Honor Societies, and is a Senior Member of IEEE. His current research interests include secure and scalable real-time IoT, applied machine learning, and applications of quantum algorithms.
\end{IEEEbiography}

\end{document}